\pdfoutput=1

\documentclass[10pt, a4paper]{article}
\usepackage[final]{lrec2026}

\usepackage{float}
\usepackage{pdfpages}
\usepackage{times}
\usepackage{latexsym}

\usepackage[T1]{fontenc}
\usepackage{amsmath}

\usepackage[utf8]{inputenc}
\usepackage{cleveref}
\usepackage{graphicx} 

\usepackage{color-edits}
\addauthor{z}{magenta}

\usepackage{microtype}

\usepackage{inconsolata}

\usepackage{multicol} 
\usepackage{cuted}
\usepackage{enumitem}

\usepackage{array}
\usepackage{lipsum}

\usepackage{soul}
\usepackage{subcaption}
\usepackage{adjustbox}

\usepackage{tikz}
\definecolor{stem}{HTML}{FFB6C1}
\colorlet{stemTransparent}{stem!20} 

\definecolor{humanities}{HTML}{ADD8E6}
\colorlet{humanitiesTransparent}{humanities!20} 

\definecolor{other}{HTML}{90EE90}
\colorlet{otherTransparent}{other!20} 

\definecolor{social_sciences}{HTML}{FFDAB9}
\colorlet{social_sciencesTransparent}{social_sciences!20} 
\usepackage{arabtex}
\usepackage{utf8}

\usepackage{booktabs}
\newcommand{\Egyptian}{EGY}
\newcommand{\Saudi}{KSA}
\newcommand{\Moroccan}{MAG}
\newcommand{\Syrian}{SYR}
\newcommand{\Emirati}{UAE}

\makeatletter

\usepackage{multirow}
\usepackage{url}

\title{\textsc{DialectalArabicMMLU}: Benchmarking Dialectal Capabilities\\in Arabic and Multilingual Language Models}

\name{Malik H. Altakrori\textsuperscript{1}\quad
Nizar Habash\textsuperscript{2,3}\quad
Abed Alhakim Freihat\textsuperscript{3}\\
\large\textbf{Younes Samih\textsuperscript{1}}\quad
\large\textbf{Kirill Chirkunov\textsuperscript{3}}\quad 
\large\textbf{Muhammed AbuOdeh\textsuperscript{3}} \\
\large\textbf{Radu Florian\textsuperscript{1}}\quad
\large\textbf{Teresa Lynn\textsuperscript{3}}\quad 
\large\textbf{Preslav Nakov\textsuperscript{3}}\quad 
\large\textbf{Alham Fikri Aji\textsuperscript{3}}
}

\address{$^1$IBM Research AI \quad $^2$New York University Abu Dhabi \\
$^3$Mohamed bin Zayed University of Artificial Intelligence \\
         Abu Dhabi,UAE \\
         \texttt{malik.altakrori@ibm.com, nizar.habash@nyu.edu, alham.fikri@mbzuai.ac.ae}
  }

\abstract{
We present \textsc{DialectalArabicMMLU}, a new benchmark for evaluating the performance of large language models (LLMs) across Arabic dialects. While recently developed Arabic and multilingual benchmarks have advanced LLM evaluation for Modern Standard Arabic (MSA), dialectal varieties remain underrepresented despite their prevalence in everyday communication. \textsc{DialectalArabicMMLU} extends the MMLU-Redux framework through manual translation and adaptation of 3K multiple-choice question–answer pairs into five major dialects (Syrian, Egyptian, Emirati, Saudi, and Moroccan), yielding a total of 15K QA pairs across 32 academic and professional domains (22K QA pairs when also including English and MSA). The benchmark enables systematic assessment of LLM reasoning and comprehension beyond MSA, supporting both task-based and linguistic analysis. We evaluate 19 open-weight Arabic and multilingual LLMs (1B--13B parameters) and report substantial performance variation across dialects, revealing persistent gaps in dialectal generalization. \textsc{DialectalArabicMMLU} provides the first unified, human-curated resource for measuring dialectal understanding in Arabic, thus promoting more inclusive evaluation and future model development.
 \\ \newline 
 }

\begin{document}
\maketitleabstract

\section{Introduction\label{sec:intro}}

The rise of large language models (LLMs) has been enabled by the development of robust evaluation benchmarks capable of assessing not only their overall performance on Natural Language Processing (NLP) tasks, but also their linguistic adaptability across languages and varieties. Many recent efforts have therefore focused on multilingual benchmarks that evaluate model capabilities beyond English. In question-answering (QA), benchmarks are often first developed in English and later extended to other languages through translation-based adaptation, as in MLQA~\citep{lewis-etal-2020-mlqa} and TyDi QA~\citep{clark-etal-2020-tydi}. Such approaches have proven effective for broadening cross-lingual evaluation coverage.

In the case of Arabic, evaluation has largely centered on Modern Standard Arabic (MSA), the standardized written variety used in formal communication. This focus overlooks the diglossic nature of Arabic, where MSA coexists with diverse regional dialects that differ substantially in morphology, syntax, lexicon, and usage~\cite{ferguson1959diglossia}. While recent Arabic benchmarks have advanced coverage and modeling for MSA, they provide limited insight into LLM behavior on \emph{Arabic dialects}, which dominate everyday communication, social media, and spoken interaction. As a result, current evaluation practices offer only a partial picture of real-world Arabic language understanding.

Here, we bridge this gap by introducing \textsc{DialectalArabicMMLU},\footnote{Available on HuggingFace: \href{https://hf.co/datasets/MBZUAI/Dialectal-Arabic-MMLU}{Dialectal-Arabic-MMLU}} a new benchmark specifically designed to evaluate LLM capabilities across five major Arabic dialects: Syrian, Egyptian, Emirati, Saudi, and Moroccan. Building upon the English-based MMLU-Redux~\cite{hendrycks:2020}, we adopt the established paradigm of adapting high-quality English QA benchmarks, but extend it to dialectal Arabic through fully manual translation.

By situating dialects, not MSA, as the primary target of evaluation, \textsc{DialectalArabicMMLU} establishes a framework for quantifying dialectal understanding, reasoning, and general knowledge in Arabic. The benchmark enables controlled comparison across parallel dialectal variants while preserving semantic equivalence. Our contributions can be summarized as follows:

\begin{itemize}

\item We introduce \textsc{DialectalArabicMMLU}, the first large-scale, human-curated benchmark specifically designed to evaluate the reasoning and comprehension capabilities of LLMs across five major Arabic dialects.

\item We develop a comprehensive dataset of more than 3K QA pairs per dialect (plus MSA and English), a total of 15K QA pairs, spanning 32 academic and professional domains, all produced and validated by native speakers to ensure linguistic fidelity and naturalness.

\item We evaluate 19 open-weight Arabic and multilingual LLMs (ranging from 1B to 13B parameters) under three experimental settings (default, oracle, and dialect identification) to systematically assess the impact of dialectal variation on model performance.

\item We conduct a detailed analysis of model behavior across dialects, revealing substantial performance disparities and highlighting the need for dialect-aware evaluation and training strategies for Arabic-enabled LLMs.

\end{itemize}

\begin{table*}[htbp]
    \centering
    \scriptsize
    \setcode{utf8}    
    \begin{tabular}{lrrrrr}
    \toprule
    \textbf{Dialect} & \multicolumn{1}{c}{\textbf{Question}}  & \multicolumn{1}{c}{\textbf{Ch. 1}} & \multicolumn{1}{c}{\textbf{Ch. 2}} & \multicolumn{1}{c}{\textbf{Ch. 3}} & \multicolumn{1}{c}{\textbf{Ch. 4}} \\
    \midrule
    MSA  & \RL{هل تتغير السمات مع التقدم في العمر؟} & \RL{لا} & \RL{تتغير قليلاً} & \RL{فقط لدى النساء} & \RL{تتغير كثيرًا} \\
    EGY  & \RL{هل الصفات الشخصية بتتغيّر لما الواحد بيكبر في السن؟} &\RL{لا} & \RL{بتتغير شوية} & \RL{بس عند الستات} & \RL{بتتغير كتير} \\
    KSA  & \RL{تتغير السمات مع التقدم في العمر؟} &\RL{لا} & \RL{تتغير شوية} & \RL{بس عند الحريم} & \RL{تتغير كثير} \\
    MAG  & \RL{واش كيتبدلو السمات (ديال الشخصية) مع التقدم فالعمر؟} & \RL{لا} & \RL{كتبدّل شوية} & \RL{غير عند العيالات} & \RL{كتتبدّل بزاف} \\
    SYR  & \RL{بتتغير ميزات البني آدم بس يكبر؟} & \RL{لأ} & \RL{بتتغير شوي} & \RL{بس عند النسوان} & \RL{بتتغير كتير} \\
    UAE  & \RL{هل تتغير الصفات الشخص مع كبر العمر؟} & \RL{لا} & \RL{تتغير شوي} & \RL{فقط عند النساء} & \RL{تتغير بشكل كبير} \\
    ENG  &  \multicolumn{1}{l}{Do traits change with age?} & No & \multicolumn{1}{l}{They change a little} & \multicolumn{1}{l}{Only for women} & \multicolumn{1}{l}{They change quite a lot} \\
    \bottomrule
    \end{tabular}
    \caption{An example from the \textbf{Human Aging} domain, showing the variation in the questions and the choices across dialects/languages. (QID 20, the correct answer is choice 2: ``\emph{They change a little.}'')}
    \label{tab:placeholder}
\end{table*}

\section{Related Work}\label{sec:related}

Several benchmarks have been introduced to enable LLM evaluation for Arabic. ArabicMMLU~\cite{koto-etal-2024-arabicmmlu} created an MMLU-like framework for MSA, drawing on real school and professional exams from around the Arab world. LAraBench~\cite{abdelali-etal-2024-larabench} collected a comprehensive suite of 61 datasets spanning 33 diverse tasks across text and speech, establishing a multi-domain, multi-task evaluation platform for Arabic. More recently, BALSAM~\cite{balsam:2025} emerged as a community-driven, unified benchmark encompassing 78 NLP tasks from 14 broad categories with over 52K examples, with curated data covering diverse domains and various Arabic dialects. However, none of the above work had dialectal Arabic evaluation of LLMs as its main focus.

Complementing these efforts, 3LM~\cite{3lm:2025} focused on bridging Arabic, STEM, and code, introducing a set of rigorous benchmarks to evaluate Arabic-enabled LLMs on scientific reasoning and programming tasks. It extended prior efforts by targeting domains that require structured problem-solving, such as mathematics, physics, and computer science, where Arabic benchmarks have been notably lacking. These initiatives, together with Arabic-centric models such as Jais~\cite{sengupta:2023:jais}, ALLaM~\cite{bari2025allam}, and Fanar~\cite{fanar:2025}
significantly advanced evaluation for MSA. Yet, they remain centered on Modern Standard Arabic and offer only minimal insight into dialectal Arabic performance.

Extensive research, benchmarks, and shared tasks have targeted dialectal Arabic \emph{identification} and \emph{translation}~\cite{bouamor-etal-2019-madar,abdul-mageed-etal-2021-nadi,abdul-mageed-etal-2022-nadi,abdul-mageed-etal-2023-nadi,abdul-mageed-etal-2024-nadi}, which was complemented by several datasets and tools~\cite{zaidan-callison-burch-2014-arabic,bouamor-etal-2014-multidialectal,salama-etal-2014-youdacc,alsarsour-etal-2018-dart,abu-kwaik-etal-2018-shami,salameh-etal-2018-fine,bouamor-etal-2018-madar,abdelali-etal-2021-qadi,baimukan-etal-2022-hierarchical}.  Unlike this work, which has focused on just two tasks, our focus here is on general dialectal understanding.

Moreover, new benchmarks have begun to address dialectal and cultural dimensions in Arabic, albeit only partially. PALM~\cite{alwajih-etal-2025-palm} introduced a year-long, human-curated dataset of over 10K instruction--response pairs covering all 22 Arab countries in both MSA and dialects, across 20 culturally salient topics. While PALM effectively exposed critical gaps in model performance on culturally grounded and dialectal instructions, its design follows an instruction-tuning paradigm rather than a multitask reasoning framework.

AraDiCE~\cite{mousi-etal-2025-aradice} offers the most comprehensive attempt so far at LLM evaluation for dialectal Arabic, contributing 45K post-edited dialectal examples across Egyptian, Syrian, and Gulf varieties of Arabic and introducing a fine-grained cultural question--answering component. While being a significant step toward dialectal evaluation by extending MMLU-style tasks to Arabic dialectal varieties, it is largely derived from machine-translation followed by post-editing. In contrast, our benchmark is entirely human-translated by native speakers of the target Arabic dialects and double-checked by other native speakers, ensuring linguistic naturalness, idiomatic precision, and cultural authenticity.

Belebele~\cite{Bandarkar:2024} extends coverage to several Arabic dialects in a controlled reading comprehension setting, but remains limited in task scope and does not assess multi-domain reasoning or knowledge in Arabic dialects.

JEEM~\cite{kadaoui:2025jeem} expands benchmarking into a multimodal setting by evaluating image captioning and visual question--answering across Jordanian, Emirati, Egyptian, and Moroccan Arabic dialects, revealing that current Arabic vision–language models (VLMs), including GPT-4V~\cite{openai:2024gpt4}, struggle with dialect-specific visual understanding and show uneven competence across dialects.

Alyah~\cite{emirati_dialect_benchmark_2026} is an Emirati-dialect benchmark that focuses on evaluation for Gulf and Emirati Arabic, offering 1,173 culturally grounded examples sourced from native speakers and covering idioms, greetings, figurative expressions, etiquette norms, and other dialect-specific linguistic features. However, it is limited to a single dialect, and is smaller in size compared to our dataset.

In contrast to the above work, we introduce a large-scale, parallel benchmark explicitly designed for understanding multiple dialectal Arabic. Unlike prior benchmarks that position dialects as secondary to MSA or cover a single dialect, our benchmark features balanced coverage across major regional varieties and relies exclusively on human-curated translation and dialectal adaptation, enabling principled evaluation of reasoning and comprehension across dialects.

\section{Data Collection and Quality Assurance}\label{sec:data_annotation}

\textsc{DialectalArabicMMLU}, is based on the translation of over 3,135 English (\textbf{ENG}) multiple-choice question--answering (MCQA) pairs into five dialects representing the geographical spread of the Arab World: Egypt~(\textbf{\Egyptian}), Morocco (\textbf{\Moroccan}), Saudi Arabia (\textbf{\Saudi}), Syria (\textbf{\Syrian}), and the United Arab Emirates (\textbf{\Emirati}). We also include \textbf{MSA} and English: a total of 21,945 MCQA pairs (see Table~\ref{tab:placeholder} for an example).

\paragraph{Dataset}
Our dataset is based on MMLU-Redux-v2 ~\cite{gema2025mmlu},\footnote{Available from HuggingFace: \href{https://huggingface.co/datasets/edinburgh-dawg/mmlu-redux-2.0}{mmlu-redux-2.0}} a high-quality re-annotated subset of MMLU. We selected 32 domains from it and translated the corresponding QA-pairs to the above five dialects. Below are the fields and the domains in each field:
\begin{itemize}
  \setlength\itemsep{0pt}
  \setlength\parskip{0pt}
  \setlength\parsep{0pt}
    \item \textbf{Humanities}: High School US History, High School World History, International Law, Moral Scenarios, Philosophy, Prehistory, Professional Law, World Religions;
    \item \textbf{Stem}: Abstract Algebra, Anatomy, Astronomy, College Computer Science, Conceptual Physics, Elementary Mathematics, High School Chemistry;
    \item \textbf{Social} \textbf{Sciences}: High School Geography, High School Macroeconomics, High School Psychology, Professional Psychology, Public Relations, Security Studies, Sociology, US Foreign Policy;
    \item \textbf{Other}: Business Ethics, Clinical Knowledge, College Medicine, Global Facts, Human Aging, Management, Marketing, Nutrition, Virology. 
\end{itemize}

The translation was carried out in two main phases: manual translation and quality assurance.

\paragraph{Manual Translation} 
The translations were outsourced to a language service provider (LSP) and carried out manually by teams of native or near-native speakers of each dialect. To ensure consistency, the process was guided by a detailed translation guidelines document (see Appendix~\ref{app:guidelines}).

The guidelines emphasized three principles:
\begin{itemize}[nosep, topsep=2pt] 
  \setlength\itemsep{0pt}
  \setlength\parskip{0pt}
  \setlength\parsep{0pt}
  
    \item \textbf{Correctness}: accurately reflect the source meaning;
    \item \textbf{Naturalness}: sound authentic in the target dialect, even if close to MSA;
    \item \textbf{Simplicity}: use concise, conversational language.
\end{itemize}

Additional instructions included avoiding over-dialectalization, using MSA terms where contextually natural, and respecting natural spelling variation in the dialectal orthography.

\noindent Each dialect team had members with three roles:
\begin{itemize}[nosep, topsep=2pt]
  \setlength\itemsep{0pt}
  \setlength\parskip{0pt}
  \setlength\parsep{0pt}
    \item \textbf{Translator (Native/Near-native)}: produced the initial translation;
    \item \textbf{Reviewer (Native)}: reviewed and accepted/rejected annotations with justification;
    \item \textbf{Adjudicator}: resolved the disagreements and ensured the final quality.
\end{itemize}

Before starting, the translators attended two training sessions focusing on the workflow and the guidelines, to clarify the expected outcome and to align across the teams.

\paragraph{Quality Assurance}
To assess the translation quality, we carried out an in-house validation step. For each dialect, we sampled 32 QA pairs from eight domains, and native speakers independently scored the translations on a Likert scale of 1--5:

\begin{enumerate}
  \setlength\itemsep{0pt}
  \setlength\parskip{0pt}
  \setlength\parsep{0pt}
    \item The translation is incorrect. 
    \item The translation is partially correct.\\ \emph{(e.g., contains some inaccuracies, contains MSA where dialectal terms exist, etc.)}.
    \item The translation is acceptable\\ \emph{(e.g., contains the original meaning but could be improved in terms of formulation, fluency)}.
    \item The translation is good\\ \emph{(but I would translate it differently)}.
    \item The translation is very good.
\end{enumerate}

This evaluation step showed that only about 5\% of the translations contained some inaccuracies. Almost all dialectical translations had average assessments between good (4) and very good (5) levels, except for the UAE dialect, which was only marginally below good (at 3.94). 

We further performed linguistic analysis, which revealed that some of the UAE translations had a bias towards the use of Saudi dialectal words instead of more common terms used in the region. For Syrian, some translators had a tendency to translate the concept rather than the original text. Additionally, there were issues with the translation of pronouns.  This evaluation effectively resulted in a second round of translation-review revision for the UAE and Syrian QA pairs, improving the quality of the final dataset. 

\begin{table}[t!]
\small
 \begin{center}
     \tabcolsep6pt
\begin{tabularx}{\columnwidth}{lrrrrrr}
\toprule
\textbf{Dial.} & \textbf{SYR} & \textbf{UAE} & \textbf{KSA} & \textbf{MSA} & \textbf{MAG} & \textbf{EGY} \\
\midrule
\textbf{SYR} & 1.0 & .45 & .55 & .51 & .34 & .52 \\
\textbf{UAE} &     & 1.0 & .59 & .61 & .37 & .51 \\
\textbf{KSA} &     &     & 1.0 & .70 & .39 & .61 \\
\textbf{MSA} &     &     &     & 1.0 & .41 & .59 \\
\textbf{MAG} &     &     &     &     & 1.0 & .37 \\
\textbf{EGY} &     &     &     &     &     & 1.0  \\
\bottomrule
\end{tabularx}
\caption{Binary Jaccard word-level similarity between the dialectal Arabic pairs.}
\label{tab:jacc_bin}
\end{center}
\end{table}

\paragraph{Similarity between Dialects}
To further understand the nature of the data, we examined the lexical variations between the dialects using Jaccard similarity. We took the frequency counts for all words in both the question and the choice columns and passed them through a preprocessing step using the CAMeL~Tools~\cite{obeid-etal-2020-camel} for Unicode character normalization, dediacritization, and whitespace tokenization. Table~\ref{tab:jacc_bin} shows the Jaccard similarity between the various dialect pairs. 

When comparing our results to those of~\newcite{salameh-etal-2018-fine}, we observe several differences. While we found MAG to be the most distant dialect from MSA, they reported it as relatively closer, and whereas they placed KSA much further from MSA, we found it to be the closest one. These discrepancies may stem from differences in data: their study used travel expressions, while our MMLU questions are more technical.

In both cases, EGY is closer to MSA than SYR. We provide the weighted Jaccard similarity measure (Table~\ref{tab:app_jacc_w}), and the Manhattan (Table~\ref{tab:app_mann_l1}) and Euclidean (Table~\ref{tab:app_Euc_l2}) distances in Appendix~\ref{app:simscores}. 

\paragraph{Dataset Statistics} Table~\ref{tab:stats_numbers} shows some dataset statistics: number of dialects, domains, questions, translations, including both average and total counts. 
Table~\ref{tab:stats_lengths} further zooms into the average length of the questions and the answers across the investigated dialects/languages in our dataset: we can see that the lengths are similar across the Arabic varieties, with English being slightly longer. We provide detailed question counts for each dialect--topic combination in Table~\ref{tab:domain_dialect_questions} in Appendix~\ref{app:longStats}.  

\section{Experimental Setup}
\label{sec:exp-setup}
In this section, we present our experimental setup, the LLMs we experiment with, and the evaluation setup.

\begin{table}[t!]
\small
    \begin{tabular}{cll}
    \toprule
    & \textbf{Characteristic} & \multicolumn{1}{c}{\bf Value}\\
    \midrule
        1. & \# of dialects &  5, plus ENG \& MSA \\        
        2. & \# of domains & 32 domains (in 4 fields)\\        
        3. & \# of Qs/domain & $\simeq$ 98.0 Qs (68--100) \\
        4. & \# of Qs/dialect & 3,135 Qs\\
        5. & \# of Qs translated & 15,675 Qs (5 * 3,135)\\
        6. & \# of Qs in total   & 21,945 Qs (7 * 3,135)\\        
    \bottomrule
    \end{tabular}
    \caption{Dataset statistics.
    \label{tab:stats_numbers}}
\end{table}

\begin{table}[t!]
\small
\centering
\tabcolsep7pt
\begin{tabular}{lccccc}
\toprule
\multirow{2}{*}{\textbf{Dialect}} & \multicolumn{2}{c}{\textbf{Questions}} & & \multicolumn{2}{c}{\textbf{Choices}} \\
\cmidrule{2-3} \cmidrule{5-6}
 &   \textbf{Chars} &  \textbf{Words} & &  \textbf{Chars} &   \textbf{Words} \\
\midrule
\textbf{\Egyptian }& 171.5 & 29.5 && 130.4 & 22.1 \\
\textbf{\Saudi}    & 170.3 & 29.8 && 129.9 & 21.8 \\
\textbf{\Moroccan} & 179.8 & 30.0 && 137.3 & 22.5 \\
\textbf{\Syrian}   & 162.7 & 27.8 && 121.0 & 20.4 \\
\textbf{\Emirati}  & 163.1 & 27.7 && 125.5 & 20.9 \\
\textbf{MSA}       & 177.5 & 30.3 && 132.3 & 22.1 \\
\textbf{ENG}       & 212.5 & 35.9 && 157.1 & 24.7 \\
\bottomrule
\end{tabular}
\caption{Average question and choice length (characters and words) across the dialects.}
\label{tab:stats_lengths}
\end{table}

\subsection{Evaluated Arabic-Enabled LLMs}
We evaluated language models of 1B to 13B parameters. These models are considered small- to medium-size compared to frontier models such as the 120B version of OpenAI's GPT-OSS. 

We based our LLM selection on~\cite{ouda2025arabic}, which provides a comprehensive survey of Arabic language models across different sizes, including both open-weight and proprietary ones. For our experiments, we restrict ourselves to open-weight LLMs only.\footnote{All selected models are publicly available on \href{www.HuggingFace.com}{HuggingFace}.}

We made sure that we included the three main Arabic-enabled LLMs that were developed in the Arab region: ALLaM~\cite{bari2025allam}, Fanar~\cite{fanar:2025}, and Jais~\cite{sengupta:2023:jais}, in addition to recent, multilingual models of comparable sizes such as Google's Gemma-3~\cite{gemma_2025} and Cohere Labs' Command R7B~\cite{alnumay2025command}. Note that these models do not distinguish between MSA and dialects; rather, all the dialects and MSA are considered as just Arabic. A list of these models, their sizes, and whether they support Arabic, English, or both, is provided in Table~\ref{tab:List_models}.   

\begin{table}[t]
\small
\tabcolsep2pt
    \centering
    \begin{tabular}{lllrcc}
\toprule
 & \textbf{Family} & \textbf{Model} & \textbf{Size} & \textbf{Ar} & \textbf{En} \\
\midrule
1 & inceptionai & jais-13b-chat & 13.0 & $\bullet$ & $\bullet$ \\
2 & google & gemma-3-12b-it & 12.2 & $\bullet$ & $\bullet$ \\
3 & MBZUAI-Paris & Nile-Chat-12B & 11.8 & $\bullet$ &   \\
4 & silma-ai & SILMA-9B-Instruct & 9.2 & $\bullet$ & $\bullet$ \\
5 & QCRI & Fanar-1-9B-Instruct & 8.8 & $\bullet$ & $\bullet$ \\
6 & CohereLabs & command-r7b-arabic & 8.0 & $\bullet$ & $\bullet$ \\
7 & CohereLabs & aya-expanse-8b & 8.0 & $\bullet$ & $\bullet$ \\
8 & tiiuae & Falcon-H1-7B-Instruct & 7.6 & $\bullet$ & $\bullet$ \\
9 & mistralai & Mistral-7B-Instruct & 7.2 & $\bullet$ & $\bullet$ \\
10 & ALLaM-AI & ALLaM-7B-Instruct & 7.0 & $\bullet$ & $\bullet$ \\
11 & Navid-AI & Yehia-7B & 7.0 & $\bullet$ & $\bullet$ \\
12 & inceptionai & jais-6p7b-chat & 6.8 & $\bullet$ & $\bullet$ \\
13 & google & gemma-3-4b-it & 4.3 & $\bullet$ & $\bullet$ \\
14 & Qwen & Qwen3-4B-Instruct* & 4.0 & $\bullet$ & $\bullet$ \\
15 & MBZUAI-Paris & Nile-Chat-4B & 3.9 & $\bullet$ &   \\
16 & UBC-NLP & NileChat-3B & 3.1 & $\bullet$ & $\bullet$ \\
17 & tiiuae & Falcon-H1-3B-Instruct & 3.1 & $\bullet$ & $\bullet$ \\
18 & inceptionai & jais-2p7b-chat & 2.7 & $\bullet$ & $\bullet$ \\
19 & stabilityai & ar-stablelm-2-chat & 1.6 & $\bullet$ &   \\
\bottomrule
\end{tabular}
    \caption{The evaluated language models. (* Based on Qwen2 language support)
    \label{tab:List_models}}
\end{table}

\begin{table*}[t!]
    \centering
    \small
    \begin{tabular}{lrccccc|c|cc}
\toprule
\textbf{Model} & \textbf{Size$\downarrow$} & \textbf{EGY} & \textbf{KSA} & \textbf{MAG} & \textbf{SYR} & \textbf{UAE} & \textbf{DA Avg} & \textbf{MSA} & \textbf{ENG}  \\
\midrule
jais-13b-chat & 13.0 & 48.0 & 48.2 & 44.7 & 45.4 & 48.5 & 47.0 & 52.0 & 55.3 \\
gemma-3-12b-it & 12.2 & 61.5 & 58.5 & 54.3 & 57.4 & 61.7 & 58.7 & 62.6 & 73.7 \\
Nile-Chat-12B & 11.8 & \textbf{61.9} & \textbf{60.6} & \textbf{55.8} & \textbf{58.9} & \textbf{62.5} & \textbf{59.9} & \textbf{63.8} & 72.8 \\
SILMA-9B-Instruct & 9.2 & 55.3 & 54.0 & 48.7 & 52.0 & 55.3 & 53.1 & 57.6 & 72.4 \\
Fanar-1-9B-Instruct & 8.8 & 58.5 & 56.6 & 53.6 & 54.6 & 58.0 & 56.2 & 61.3 & 70.4 \\
command-r7b-arabic & 8.0 & 53.5 & 52.8 & 50.2 & 52.2 & 55.0 & 52.7 & 57.7 & 67.4 \\
aya-expanse-8b & 8.0 & 51.8 & 50.1 & 47.2 & 49.2 & 52.1 & 50.1 & 54.0 & 63.3 \\
Falcon-H1-7B-Instruct & 7.6 & 59.1 & 58.1 & 52.6 & 55.8 & 60.2 & 57.2 & 62.4 & \textbf{76.5} \\
Mistral-7B-Instruct & 7.2 & 33.9 & 35.2 & 34.0 & 33.3 & 36.2 & 34.5 & 38.0 & 62.8 \\
ALLaM-7B-Instruct & 7.0 & 56.6 & 56.2 & 53.4 & 55.3 & 58.2 & 56.0 & 60.3 & 66.7 \\
Yehia-7B & 7.0 & 53.7 & 53.6 & 50.5 & 52.9 & 55.3 & 53.2 & 58.5 & 62.5 \\
jais-6p7b-chat & 6.8 & 42.8 & 44.6 & 40.2 & 41.5 & 45.3 & 42.9 & 48.2 & 52.9 \\
gemma-3-4b-it & 4.3 & 41.0 & 40.5 & 36.4 & 38.0 & 43.3 & 39.8 & 44.1 & 54.6 \\
Qwen3-4B-Instruct & 4.0 & 28.7 & 27.1 & 26.7 & 26.6 & 28.8 & 27.6 & 31.2 & 65.5 \\
Nile-Chat-4B & 3.9 & 48.3 & 47.0 & 42.4 & 45.3 & 47.8 & 46.2 & 49.5 & 59.3 \\
Falcon-H1-3B-Instruct & 3.1 & 46.1 & 44.7 & 41.7 & 43.1 & 46.2 & 44.3 & 48.4 & 67.9 \\
NileChat-3B & 3.1 & 54.3 & 51.8 & 52.8 & 51.0 & 53.7 & 52.7 & 55.6 & 64.3 \\
jais-2p7b-chat & 2.7 & 38.2 & 40.4 & 34.4 & 37.7 & 41.5 & 38.4 & 43.4 & 47.1 \\
ar-stablelm-2-chat & 1.6 & 36.4 & 36.5 & 35.5 & 36.1 & 36.4 & 36.2 & 37.3 & 38.3 \\
\midrule
\textbf{Average} & 6.8 & 48.9 & 48.2 & 45.0 & 46.6 & 49.8 & 47.7 & 51.9 & 62.8 \\
\midrule
\textit{Jais-2-8B-Chat} & 8.1  & 56.9 & 55.7 & 53.0 & 54.1 & 56.8 & 55.3 & 59.0 & 65.3 \\
\textit{Jais-2-70B-Chat}              & 72.4 & 68.0 & 68.0 & 65.0 & 66.0 & 68.8 & 67.2 & 70.7 & 76.2 \\
\bottomrule 
\end{tabular}
\caption{Accuracy scores for the default \textsc{DialectalArabicMMLU} setting. (Average of 5 runs for the 32 different topics. Random chance = $\frac{1}{4}$. Size$\downarrow$: Sorted in descending order. \textbf{Bold}: Maximum per column.)\label{tab:mmlu_default}}
\end{table*}

\subsection{QA Evaluation Setup}\label{subsec:lmEval} 
For our experiments, we adopted the LM-Eval-Harness framework~\cite{eval-harness}, which is a community-supported tool that contains a suite of evaluation tasks to measure the performance of LLMs. We developed custom evaluation modules based on the original MMLU configuration and extended it for our scenarios: 

\begin{itemize}
    \item \textbf{Default Setting} This setting preserves the original MMLU prompt without any dialectal cues. It varies, however, based on the evaluated domain, e.g.,~for the \textit{Abstract Algebra} domain, the prompt will be \textit{``The following are multiple choice questions (with answers) about abstract algebra.''} followed by the multiple choices, and concluded with \textit{``Answer:''} 

\item \textbf{Oracle Setting} This setting introduces explicit dialectal conditioning by specifying the dialect of the question as part of the prompt. As a result, the prompt is modified based on the dialects as well. For the same domain, \textit{Abstract Algebra}, the first part of the prompt in the oracle setting will be \textit{``The following are multiple choice questions (with answers) about abstract algebra in the \textul{Egyptian dialect}.''} 

\item \textbf{Dialect Identification}
The {Dialect Identification} setting asks the model to infer the dialect of the input: each of our five dialects or MSA. Here, the topic is irrelevant and, as a result, we have a fixed prompt: \textit{``The following are multiple-choice questions (with answers) for Arabic dialect identification.''} 
\end{itemize}

All tasks follow a multiple-choice format to align with MMLU. We use log-likelihood evaluation, appending each answer option to the prompt and selecting the one with the highest log-likelihood as the model's prediction. Then, the accuracy is determined by exact match with the gold answer. Each experiment is repeated five times, and we report the average accuracy across runs and across the 32 topics per dialect/language (unless stated otherwise). This setup ensures transparency, comparability, and controlled analysis of model sensitivity to dialectal variation.

\section{Experimental Results and Analysis}\label{sec:experiments}
In this section, we discuss the experiments and the analysis, organized around three key questions:

\subsection{MSA vs. DA in QA Performance}
\label{subsec:default_mmlu}

\paragraph{How do LLMs perform on Question--Answering tasks in MSA compared to dialectal Arabic?}
Table~\ref{tab:mmlu_default} shows the accuracy for various LLMs when evaluated on the default \textsc{DialectalArabicMMLU} setting for QA. For each model--dialect pair, we report the average accuracy over the 32 domains, with the experiment repeated five times to ensure stability and robustness.

\begin{table*}[htbp]
    \centering
    \small
    \tabcolsep6pt
\begin{tabular}{lrrrrrrr|r|r}
\toprule
\textbf{Model} & \textbf{Size$\downarrow$} & \textbf{EGY} & \textbf{KSA} & \textbf{MAG} & \textbf{SYR} & \textbf{UAE} & \textbf{MSA} & \textbf{DA Avg} & \textbf{Avg All} \\
\midrule
jais-13b-chat & \textbf{13.0} & 40.0 & 24.1 & 10.1 & 16.8 & 13.4 & 26.7 & 20.9\phantom{-.-} & 21.8\phantom{-.-} \\
gemma-3-12b-it & 12.2 & 36.0 & 9.5 & 64.0 & \textbf{71.0} & 4.8 & 83.9 & 37.1\phantom{-.-} & 44.9\phantom{-.-} \\
Nile-Chat-12B & 11.8 & 30.5 & 0.6 & 22.6 & 19.9 & 0.5 & 82.7 & 14.8\phantom{-.-} & 26.1\phantom{-.-} \\
SILMA-9B-Instruct & 9.2 & 54.8 & 13.1 & 59.4 & 29.6 & 10.2 & 58.5 & 33.4\phantom{-.-} & 37.6\phantom{-.-} \\
Fanar-1-9B-Instruct & 8.8 & \textbf{84.2} & 9.3 & 45.5 & 8.6 & 0.9 & 58.8 & 29.7\phantom{-.-} & 34.5\phantom{-.-} \\
command-r7b-arabic & 8.0 & 23.4 & 6.4 & 28.8 & 17.5 & 0.5 & 87.2 & 15.3\phantom{-.-} & 27.3\phantom{-.-} \\
aya-expanse-8b & 8.0 & 38.2 & 3.9 & 17.2 & 3.7 & 1.4 & 59.0 & 12.9\phantom{-.-} & 20.6\phantom{-.-} \\
Falcon-H1-7B-Instruct & 7.6 & 63.5 & 2.4 & 24.3 & 6.4 & 0.4 & 74.8 & 19.4\phantom{-.-} & 28.6\phantom{-.-} \\
Mistral-7B-Instruct & 7.2 & 37.2 & 0.9 & 1.4 & 7.0 & 0.3 & 60.5 & 9.4\phantom{-.-} & 17.9\phantom{-.-} \\
ALLaM-7B-Instruct & 7.0 & 42.4 & 23.0 & 55.5 & 22.0 & 4.0 & \textbf{95.4} & 29.4\phantom{-.-} & 40.4\phantom{-.-} \\
Yehia-7B & 7.0 & 15.6 & 11.9 & 29.3 & 5.4 & 3.8 & 94.9 & 13.2\phantom{-.-} & 26.8\phantom{-.-} \\
jais-6p7b-chat & 6.8 & 19.5 & 3.4 & 5.6 & 6.8 & 7.2 & 68.6 & 8.5\phantom{-.-} & 18.5\phantom{-.-} \\
gemma-3-4b-it & 4.3 & 30.5 & 19.8 & 37.6 & 21.0 & 11.5 & 12.2 & 24.1\phantom{-.-} & 22.1\phantom{-.-} \\
Qwen3-4B-Instruct & 4.0 & 47.9 & 10.8 & 13.8 & 12.0 & 10.4 & 21.1 & 19.0\phantom{-.-} & 19.3\phantom{-.-} \\
Nile-Chat-4B & 3.9 & 48.4 & 4.8 & 11.4 & 12.9 & 13.2 & 11.5 & 18.1\phantom{-.-} & 17.0\phantom{-.-} \\
Falcon-H1-3B-Instruct & 3.1 & 49.0 & 3.0 & 3.9 & 4.6 & 2.0 & 60.0 & 12.5\phantom{-.-} & 20.4\phantom{-.-} \\
NileChat-3B & 3.1 & 24.8 & 2.3 & 7.7 & 11.8 & 0.2 & 80.5 & 9.4\phantom{-.-} & 21.2\phantom{-.-} \\
jais-2p7b-chat & 2.7 & 11.1 & 2.0 & 9.3 & 5.7 & 19.6 & 60.4 & 9.5\phantom{-.-} & 18.0\phantom{-.-} \\
ar-stablelm-2-chat & 1.6 & 11.0 & 4.8 & 31.5 & 10.0 & 5.3 & 46.5 & 12.5\phantom{-.-} & 18.2\phantom{-.-} \\
\midrule
\textbf{Average} & 6.8 & 37.3 & 8.2 & 25.2 & 15.4 & 5.8 & 60.2 & 18.4\phantom{-.-} & 25.3\phantom{-.-} \\
\midrule
\midrule
CAMeL~Tools-DID$_{country}$ & -- & 53.9 & 10.0 & 70.2 & 23.7 & 0.0* & 73.9 & 31.6\phantom{-.-} & 37.4\phantom{-.-}\\
CAMeL~Tools-DID$_{aligned}$ & -- & 57.4 & \textbf{31.0} & \textbf{79.4} & 64.4 & \textbf{29.6} & 73.9 & \textbf{52.4}\phantom{-.-} & \textbf{56.3}\phantom{-.-}\\
\bottomrule
\end{tabular}
\caption{
Recall scores for the Dialect Identification setting. (The average of 5 runs for the 32 different topics. The random chance is $\frac{1}{6}$ $\simeq$ 16.7. * No labels for UAE cities.) 
    \label{tab:mmlu_Dial_ID}}
\end{table*}

We can see that our newly developed dialectal Arabic evaluation dataset is effective for testing the dialectal capabilities of LLMs by highlighting the performance gap for English vs. MSA and dialects, which is easy to see given the parallel nature of the questions and the answers in the dataset.

We further see that the performance consistently declines across all dialects compared to MSA and English, and this trend holds consistently across all Arabic-enabled LLMs we evaluated. 

Finally, although comparing and ranking individual models is tempting, we deliberately refrain from doing so. Instead, we focus on the \emph{average performance across all models}.\footnote{Jais-2 was released after we performed the experiments and the detailed analysis. We included it in Table~\ref{tab:mmlu_default} for comparison, but we exclude it from the average and we do not include it in any other tables or figures.}
 We argue that this offers a more holistic perspective on the current state of the art and may yield deeper insights than analyzing individual models in isolation. One observation supporting this view is that some larger models, both Arabic-centric and multilingual, perform worse than smaller ones. Understanding this would require investigating each model's training process, including the base model (if any) and the datasets. 
Given the limited availability of such information, a fine-grained comparative analysis is impractical, which motivates our emphasis on studying the aggregate trends instead.

\subsection{DA Identification vs. QA Performance}
\label{subsec:dial_Id}

\paragraph{To what extent does a model’s proficiency in recognizing dialectal Arabic correlate with its Question--Answering performance for the same dialect?}
We evaluate the performance of the LLMs as per the setup described in Section~\ref{subsec:lmEval}. To establish a baseline, we use the CAMeL~Tools \textbf{D}ialect \textbf{ID}entification (DID) tool~\cite{obeid-etal-2020-camel}, which classifies Arabic texts into 26 dialects: MSA or one of 25 cities in 15 Arab countries; the tool can return a city, a country, or a region. 
In our experiments, we used CAMeL~Tools DID with two configurations: DID$_{country}$ and DID$_{aligned}$. For DID$_{country}$, we used the tool out-of-the-box, where the only post-processing we did was to remap the labels, e.g., \emph{Syria} is mapped to \emph{SYR}. 
For DID$_{aligned}$, we aligned the CAMeL~Tools country labels to our ones. This alignment is based on apriori geographical and dialectal groupings: 
\textbf{EGY} (Egypt, Sudan),
\textbf{KSA} (Saudi Arabia, Yemen, Baghdad/Iraq),
\textbf{MAG} (Morocco, Algeria, Tunisia, Libya),
\textbf{SYR} (Syria, Jordan, Lebanon, Palestine, Mosul/Iraq), and
\textbf{UAE} (Qatar, Oman, Basra/Iraq).

Table~\ref{tab:mmlu_Dial_ID} shows the accuracy for predicting the dialect of the question. Several observations can be made from these results. 
First, there is a huge difference between the average performance on MSA and the Arabic dialects, and many models perform worse than random. 

Moreover, CAMeL~Tools, an off-the-shelf tool with minimal alignment, achieved the best identification accuracy for three of five dialects, as well as on both the dialectal and overall averages.

\begin{table*}[t!]
\centering    
    \small
    \begin{tabular}{lrrrrr rr}
\toprule
\textbf{Model} & \textbf{EGY} & \textbf{KSA} & \textbf{MAG} & \textbf{SYR} & \textbf{UAE} & \textbf{MSA} & \textbf{ENG} \\
\midrule
jais-13b-chat & -1.0 & -1.1 & -0.6 & -0.8 & -1.6 & 0.0 & 0.0 \\
gemma-3-12b-it & -9.0 & -5.0 & -11.4 & -16.3 & -7.8 & 0.3 & 0.1 \\
Nile-Chat-12B & -1.6 & -3.1 & -1.7 & -2.3 & -3.0 & 0.1 & 0.0 \\
SILMA-9B-Instruct & -2.1 & -1.3 & -1.1 & -0.5 & -0.9 & 0.0 & 0.0 \\
Fanar-1-9B-Instruct & -2.7 & -1.1 & -2.6 & -1.0 & -1.5 & -0.1 & -0.1 \\
command-r7b-arabic & -0.7 & -0.5 & -1.4 & -0.3 & -1.1 & 0.0 & 0.1 \\
aya-expanse-8b & -1.7 & -0.8 & -0.6 & -0.8 & -2.2 & -0.1 & 0.0 \\
Falcon-H1-7B-Instruct & -3.8 & -2.7 & -1.3 & -3.0 & -2.8 & -0.2 & -0.1 \\
Mistral-7B-Instruct & -1.5 & -1.4 & -0.4 & 0.4 & -1.3 & 0.1 & -0.1 \\
ALLaM-7B-Instruct & -2.1 & -2.1 & -3.0 & -2.9 & -1.8 & 0.1 & 0.2 \\
Yehia-7B & -1.1 & -0.5 & -2.0 & -0.6 & -1.4 & 0.0 & 0.0 \\
jais-6p7b-chat & -0.3 & -1.9 & -1.1 & -1.2 & -3.7 & 0.0 & 0.0 \\
gemma-3-4b-it & -7.0 & -9.4 & -2.6 & -8.4 & -9.7 & 0.0 & 0.4 \\
Qwen3-4B-Instruct & 3.9 & 1.6 & 0.3 & 2.2 & 2.1 & -0.2 & 0.0 \\
Nile-Chat-4B & -2.5 & -1.4 & -2.1 & -1.4 & -2.2 & 0.0 & 0.2 \\
Falcon-H1-3B-Instruct & -2.8 & -2.2 & -2.1 & -1.5 & -2.3 & -0.1 & 0.1 \\
NileChat-3B & -1.7 & -0.3 & -2.1 & -1.1 & -1.6 & 0.0 & 0.1 \\
jais-2p7b-chat & 0.2 & -0.2 & 0.0 & 0.1 & -2.0 & 0.0 & 0.0 \\
ar-stablelm-2-chat & -0.2 & 0.3 & -0.5 & -1.2 & 0.4 & 0.0 & 0.0 \\
\midrule
\textbf{Average} & -2.0 & -1.7 & -1.9 & -2.1 & -2.3 & -0.0 & 0.0 \\
\bottomrule
\end{tabular}
    \caption{Difference between the accuracy scores for Oracle $-$ Default. 
    \label{tab:Oracle_default}}
\end{table*}

Finally, we emphasize the high risk resulting from combining MSA and Arabic dialects as one language. As demonstrated in Table~\ref{tab:mmlu_Dial_ID}, command-r7b-arabic which scored among the highest MSA and total average scores, performs extremely poorly on UAE and KSA dialects. 

To answer the question at the beginning of this section, we conducted a Pearson correlation analysis on the average dialectal performance, and the score of each dialect separately\footnote{See Fig.~\ref{fig:app_correlations} in Appendix~\ref{app:correlations}.}. We observed a moderate positive correlation between the MCQ and the dialect ID tasks as indicated by a Pearson correlation $r = 0.431$, which, however, is not statistically significant ($p=0.07$). Similarly, the correlation was not statistically significant for EGY, KSA, SYR and UAE with $p = 0.18$, $p = 0.79$, $p = 0.12$, and $p = 0.07$, respectively. The difference was statistically significant only for MAG with $p = 0.04$, for a moderate positive correlation of $r = 0.483$. 

We investigated this behavior further using the Oracle setting explained in Section~\ref{subsec:lmEval}, where we infused the prompt with extra information about the dialect. Based on the results above, our intuition is that a model that cannot identify the dialect will not benefit from being told what that dialect is. 

The results are shown in Table~\ref{tab:Oracle_default}, where we can see the difference in accuracy between the Oracle setting, where we inject the dialect ID in the prompt and the default setting, where the prompt contains no explicit information about the dialect.\footnote{The exact results are in Table~\ref{tab:mmlu_oracle} in Appendix \ref{app:add-results}.} 

We did not expect the added dialect label to improve performance, as it provides little useful signal to the LLMs. Indeed, the oracle setting led to a statistically significant drop across all dialects. While the link between dialect identification and QA remains unclear, explicitly priming the model with the dialect label consistently degraded performance.

\begin{table}[t!]
    \centering
    \small
    \tabcolsep2pt
    \begin{tabular}{lc|cc|c}
    \toprule
\textbf{Model} & \textbf{MSA} & \textbf{MADLAD} & \textbf{Google} & \textbf{ENG} \\
\midrule
jais-13b-chat & 52.0 & 47.6 & 50.5 & 55.3 \\
gemma-3-12b-it & 62.6 & 60.8 & 66.8 & 73.7 \\
Nile-Chat-12B & 63.8 & 60.3 & 65.5 & 72.8 \\
SILMA-9B-Instruct & 57.6 & 60.0 & 64.6 & 72.4 \\
Fanar-1-9B-Instruct & 61.3 & 58.3 & 63.1 & 70.4 \\
command-r7b-arabic & 57.7 & 57.0 & 61.8 & 67.4 \\
aya-expanse-8b & 54.0 & 54.5 & 57.2 & 63.3 \\
Falcon-H1-7B-Instruct & 62.4 & 63.5 & 69.1 & 76.5 \\
Mistral-7B-Instruct & 38.0 & 52.0 & 56.3 & 62.8 \\
ALLaM-7B-Instruct & 60.3 & 55.6 & 61.1 & 66.7 \\
Yehia-7B & 58.5 & 52.4 & 56.5 & 62.5 \\
jais-6p7b-chat & 48.2 & 46.2 & 48.3 & 52.9 \\
gemma-3-4b-it & 44.1 & 48.5 & 50.4 & 54.6 \\
Qwen3-4B-Instruct & 31.2 & 54.3 & 59.4 & 65.5 \\
Nile-Chat-4B & 49.5 & 49.1 & 54.2 & 59.3 \\
Falcon-H1-3B-Instruct & 48.4 & 55.5 & 61.5 & 67.9 \\
NileChat-3B & 55.6 & 54.2 & 59.0 & 64.3 \\
jais-2p7b-chat & 43.4 & 41.7 & 44.4 & 47.1 \\
ar-stablelm-2-chat & 37.3 & 37.0 & 37.4 & 38.3 \\
\midrule
\textbf{Average} & \textbf{51.9} & \textbf{53.1} & \textbf{57.2} & \textbf{62.8} \\
\bottomrule
    \end{tabular}
    \caption{Experiments with MSA and English input, as well as for two machine-translations of the MSA input to English: MADLAD and Google.
    \label{tab:t2EN}}
\end{table}

\begin{table*}[t!]
    \centering
    \small
    \tabcolsep2pt
    \begin{tabular}{lrrrrr|r||rrrrr|r}
\toprule
&\multicolumn{6}{c||}{\textbf{Dialectal Q\&As Translated to English}} &\multicolumn{6}{c}{\textbf{Dialectal Q\&As Translated to MSA}* } \\
\textbf{Model}& \textbf{EGY} & \textbf{KSA} & \textbf{MAG} & \textbf{SYR} & \textbf{UAE} & \textbf{Avg} & \textbf{EGY} & \textbf{KSA} & \textbf{MAG} & \textbf{SYR} & \textbf{UAE} & \textbf{Avg} \\
\midrule
jais-13b-chat & -0.1 & 0.3 & 0.8 & 1.4 & 0.9 & 0.6 & 1.4 & 0.8 & 1.4 & 2.4 & -0.2 & 1.1 \\
gemma-3-12b-it & 2.2 & 3.3 & 2.8 & 3.3 & 1.7 & 2.6 & -2.4 & 0.4 & -0.5 & -0.5 & -2.4 & -1.1 \\
Nile-Chat-12B & 1.2 & 1.2 & 1.2 & 1.5 & 0.0 & 1.0 & -2.4 & -1.8 & -0.4 & -1.0 & -2.9 & -1.7 \\
SILMA-9B-Instruct & 6.7 & 6.0 & 7.1 & 6.7 & 6.8 & 6.6 & -1.6 & -1.0 & 2.0 & 0.4 & -0.5 & -0.2 \\
Fanar-1-9B-Instruct & 1.4 & 1.4 & 0.7 & 3.0 & 2.0 & 1.8 & -1.3 & 0.2 & -1.0 & 1.0 & -0.3 & -0.2 \\
command-r7b-arabic & 5.1 & 4.3 & 3.5 & 3.1 & 3.8 & 4.0 & 0.4 & 1.0 & 0.3 & 1.5 & -0.8 & 0.5 \\
aya-expanse-8b & 4.5 & 4.3 & 4.2 & 4.2 & 4.0 & 4.2 & -1.2 & 0.6 & 2.5 & 0.8 & -0.4 & 0.5 \\
Falcon-H1-7B-Instruct & 6.7 & 5.7 & 6.2 & 7.2 & 5.1 & 6.1 & -2.0 & -0.8 & 0.5 & 0.0 & -2.2 & -0.9 \\
Mistral-7B-Instruct & 19.8 & 18.0 & 15.6 & 18.2 & 17.7 & 17.9 & 1.8 & -0.2 & 0.8 & 1.8 & -0.2 & 0.8 \\
ALLaM-7B-Instruct & 1.5 & 1.3 & -0.4 & 2.8 & 1.5 & 1.3 & -0.3 & -0.4 & -0.6 & -0.3 & -0.1 & -0.4 \\
Yehia-7B & 1.0 & -0.4 & -1.3 & -0.2 & 0.7 & 0.0 & 0.5 & -0.1 & 1.3 & -0.4 & -0.6 & 0.1 \\
jais-6p7b-chat & 3.1 & 1.3 & 2.2 & 2.9 & 1.2 & 2.1 & 3.2 & 0.7 & 3.9 & 2.5 & 0.1 & 2.1 \\
gemma-3-4b-it & 7.2 & 8.2 & 9.0 & 9.6 & 5.5 & 8.0 & 1.2 & 1.8 & 3.4 & 2.9 & -0.8 & 1.8 \\
Qwen3-4B-Instruct & 27.8 & 28.4 & 24.4 & 28.4 & 28.1 & 27.4 & 1.0 & 2.3 & 1.8 & 2.2 & 1.1 & 1.7 \\
Nile-Chat-4B & 2.6 & 4.5 & 4.9 & 6.1 & 5.4 & 4.7 & -1.4 & 0.0 & 1.5 & 0.5 & 0.0 & 0.1 \\
Falcon-H1-3B-Instruct & 12.8 & 12.9 & 11.4 & 13.8 & 12.5 & 12.7 & -0.4 & 1.3 & 1.6 & 1.7 & 0.6 & 1.0 \\
NileChat-3B & 2.3 & 2.8 & -2.0 & 3.9 & 3.4 & 2.1 & -1.9 & -0.3 & -3.7 & 0.4 & -1.2 & -1.3 \\
jais-2p7b-chat & 4.4 & 1.9 & 5.8 & 4.4 & 1.3 & 3.6 & 2.9 & 1.3 & 5.2 & 1.4 & 0.1 & 2.2 \\
ar-stablelm-2-chat & -0.5 & 1.1 & -1.5 & 1.3 & 1.1 & 0.3 & 0.4 & 0.2 & 0.0 & -0.8 & 0.3 & 0.0 \\
\midrule
\textbf{Average \tiny (SD)} & 5.8 \tiny (7.2) & 5.6 \tiny (7.2) & 5.0 \tiny (6.6) & 6.4 \tiny (7.0) & 5.4 \tiny (7.0) & 5.6 \tiny (6.9) & -0.1 \tiny (1.7) & 0.3 \tiny (1.0) & 1.1 \tiny (2.0) & 0.9 \tiny (1.2) & -0.6 \tiny (1.0) & 0.3 \tiny (1.1) \\
\bottomrule
    \end{tabular}
    \caption{Change in accuracy when using the translated inputs instead of the original inputs. The translation is performed using Google's translation API (* The dialectal Arabic questions were translated to English first and then to MSA).}\label{tab:translations_Goog}
\end{table*}
\subsection{Improving Dialectal QA through MT}
\paragraph{Can machine translation mitigate data scarcity in dialectal QA?}
In this experiment, we investigate whether translating the dialectal questions to English (or to MSA) can help language models perform better on the QA task.

\paragraph{Choosing a translation model.} To perform the translation, we experimented with two machine translation tools: (\emph{i})~Google Translate (paid) API, which is a commercial translating tool, and (\emph{ii})~Google's MADLAD--400~\cite{kudugunta2023madlad400}, which is a 7B/10B parameters, open-weight translation model. 

\paragraph{Translating the MSA question to English}
We translated the MSA input into English using the two tools above and evaluated the same set of Arabic-enabled LLMs. Table~\ref{tab:t2EN} reports the performance on the original \textbf{MSA} input, on the \textbf{MADLAD}-400 and \textbf{Google} translations, and on the original \textbf{ENG} questions. We can see that translating MSA into English improves the average performance compared to using the original MSA input for both translation models, although the results still fall short of those obtained when using the original English input. A paired t-test indicates no statistically significant difference between the original MSA and the MADLAD-translated English {\scriptsize(T-stat=0.78, P-value=0.45)}. In contrast, the following four comparisons are statistically significant:

\begin{itemize}
    \item  MSA    vs. Google {\scriptsize(T-stat$=$3.30, P-value$\leq$.000)}
    \item  MADLAD vs. Google {\scriptsize(T-stat$=$-12.12, P-value$\leq$.000)}
    \item  MADLAD vs. ENG    {\scriptsize(T-stat$=$-14.51, P-value$\leq$.000)}
    \item  Google vs. ENG    {\scriptsize(T-stat$=$-15.15, P-value$\leq$.000)}.
\end{itemize}

Based on these results, we decided to use the Google API translations when doing the evaluation. We provide the MADLAD results in Appendix~\ref{app:add-results} (see Table~\ref{tab:translations_MAD}) for reproducibility given that the technical details of the current Google API translation model is not public and it is not clear if/when this particular model would be replaced. 

\paragraph{Translating the dialectal Arabic input to English \& to MSA}

Table~\ref{tab:translations_Goog} shows the effect of translating the dialectal Arabic input to English and to MSA (using English as a pivot). The main observation is that translating to English, on average, yields performance gains when compared to using the original dialectal Arabic input (statistically significant: {\scriptsize(T-stat$=$-3.54, P-value=0.002)}. Most of this gain is driven by two multilingual models, namely \textit{Mistral-7B-Instruct} and \textit{Qwen3-4B-Instruct} with an increase of 17.9 and 27.4 points absolute, respectively. 

In contrast, when translating to MSA, nearly all the performance gains vanish to the point where the average difference between using the original and the translated questions drops from 5.6 to 0.3 points absolute, resulting in a statistically insignificant difference {\scriptsize(T-stat$=$-1.22, P-value=0.24)} in performance between using the original dialectal Arabic questions and translating them to MSA. One potential explanation is that translation errors that occur when translating to English cause more errors when translating to MSA. This behavior is consistent across all Arabic dialects as can be inferred from the average scores and their standard deviation values. 

\section{Conclusion and Future Work}
\label{sec:conclusion}

We introduced \textsc{DialectalArabicMMLU}, a new benchmark for evaluating large language models (LLMs) across major Arabic dialects. Our work addresses a persistent gap in current Arabic NLP evaluation, which has largely focused on Modern Standard Arabic (MSA) while neglecting the linguistic diversity of real-world Arabic usage. 

\textsc{DialectalArabicMMLU} extends the MMLU-Redux framework through high-quality, human-curated translations of more than 3K question--answer pairs into five dialects (Syrian, Egyptian, Emirati, Saudi, and Moroccan) resulting in a corpus of over 15K (21K when including MSA and English) multiple-choice QA instances spanning 32 academic and professional domains.

Through comprehensive experiments with nineteen open-weight Arabic and multilingual LLMs, we demonstrated that model performance drops substantially across dialects compared to MSA and English. 
We further showed that explicit dialect conditioning does not consistently improve the performance and that a model's ability to identify a dialect only moderately correlates with its ability to reason in that dialect. 
These findings underscore the need for dedicated resources and training strategies that explicitly target dialectal Arabic.

In future work, we aim to expand coverage to additional Arabic dialects and domains, including low-resource varieties and specialized professional contexts. 
Second, we want to add auxiliary tasks that probe lexical, syntactic, and pragmatic understanding in dialects. 
Finally, we envision the benchmark serving as a foundation for fine-tuning and adaptation, encouraging the development of LLMs that can reason and communicate effectively across the full spectrum of Arabic varieties.

\section{Acknowledgments}
This work was conducted as part of the \textbf{IBM–MBZUAI AI Center of Excellence}. 

The authors gratefully acknowledge the contributions of 
Alya Almsouti, 
Hamad Alshehhi, 
Mohammad Anwar, 
Samar Mohamed Magdy, and 
Tareq Almsouti 
for their assistance with the data annotation process.

\section*{Ethics and Broader Impact}

We followed ethical research and data management practices at all stages of data collection, translation, and validation. 
All question--answer pairs originate from publicly available and educational sources contained in MMLU-Redux, and carry no personal or sensitive information. All dialectal translations were produced by qualified speakers through a paid language service provider under informed consent, ensuring fair compensation and professional oversight. No personally identifiable or user-generated content was collected.

As dialectal Arabic is inherently diverse, we recognize the potential for bias arising from regional, social, or stylistic variation in translation. To minimize this, all data underwent multi-stage review by annotators from different dialectal backgrounds, with guidelines emphasizing neutrality, inclusivity, and linguistic authenticity. 
Nevertheless, residual biases reflecting the translators' linguistic preferences or educational backgrounds may persist.

The benchmark is intended exclusively for research and educational purposes. By providing an open, transparent, and reproducible evaluation framework, we aim to promote progress in Arabic NLP and raise awareness of dialectal variation as a key dimension of Arabic LLMs. We encourage responsible use of our dataset, with careful consideration of the potential downstream impact of Arabic LLM evaluation and deployment.

\section*{Limitations}

While \textsc{DialectalArabicMMLU} represents an important step toward evaluating LLMs across Arabic dialects, several limitations should be acknowledged. First, despite our focus on five major dialects (Syrian, Egyptian, Emirati, Saudi, and Moroccan), the benchmark does not yet cover the full spectrum of dialectal variation across the Arabic-speaking world. Numerous regional, urban–rural, and sociolectal sub-varieties exist within each dialect group, reflecting differences in geography, age, education, and social context, which our dataset does not explicitly represent. 

Second, dialectal Arabic lacks standardized orthography, which introduces inherent variability in spelling and transcription. Although all items were manually curated by native speakers and validated for linguistic fidelity, residual inconsistencies may still affect model evaluation. Similarly, human translation and adjudication introduce subjective judgment, which, while mitigated through multi-stage review, cannot be entirely eliminated.

Third, our experiments are limited to open-weight models of moderate size (between 1B and 13B parameters). Results for larger proprietary models, which are often stronger on multilingual tasks, remain to be explored. Finally, as our benchmark is derived from question--answering tasks, it captures only a subset of dialectal capabilities; future work should complement it with generative, conversational, and multimodal evaluations.

These limitations provide avenues for future refinement and broader representational coverage.

\section{Bibliographical References}
\bibliographystyle{lrec2026-natbib}
\bibliography{main}

\newpage
\onecolumn
\appendix 

\section{Translation Guidelines}
\label{app:guidelines}

Below are the guidelines provided to the translation team. Two online training sessions were held prior to commencing the task, to ensure that that the guidelines were clear and that all translators understood what was expected from them. The training session focused on Egyptian Arabic as an example as it was the commonly understood dialect amongst the speakers of the various dialects. 

\begin{itemize}
    \item  The translation requests are presented in an Excel sheet:\\
   Column A - English source text\\
   Column B - MSA translation (this column will be hidden)\\
   Column C - Dialect translation \\
   Column D - Reject/Accept (leave blank if translation request is accepted)\\
   Column E - Justification (If Column D is “Reject”)\\

\item The translation task will be completed sheet by sheet by two translators. One who translates and one who reviews the translation.
\item If there is a problem with the source text, the reviewer can refuse to do the translation by providing a justification for his/her rejection in Column E.
\item The rejected lines are reviewed by the team leader who decides what to do with the rejected translations. 
\item In addition, the team leader reviews the accepted translations randomly to check the quality of the review process.

\end{itemize}

\begin{flushleft}
Please pay attention to the following instructions:
\end{flushleft}

\begin{enumerate}
    \item Only work on a sheet if you are confident in your translation capabilities for the specific domain of those questions.
    \item Do not translate in cases where you doubt your understanding of the question/answers. In such cases, simply move to the next line to avoid wasting time.
    \item The English questions/answers should be used as the main reference for translation. However, it is acceptable to use the MSA translation (unhide Column B) as an additional reference when needed.
    \item Support tools: It is acceptable to use dictionaries or term bases to find the most suitable term for a word in a sentence. However, automatic translation is strictly forbidden as these translated texts will be part of a scientific study that cannot be influenced by translation technology.
    \item When translating, be simple, concise, and concrete.

    \item The final translation should read like a question-answer conversation with a friend.
    \setcode{utf8} For example:\\
    \textbf{En}: \emph{Can you help me find my bag? Of course.}\\
    \textbf{MSA}: \RL{هل يمكنك مساعدتي في ايجاد امتعتي؟ بالطبع}\\
    \textbf{Egy DA}:\RL{ساعدني القى شنطتي/  ممكن تساعدني القى شنطتي؟ اكيد}

\item MSA terms may be used, depending on the context. However, do not try to translate domain-specific words into a dialect if the context does not allow for it.

\item In some cases, the resulting translation may seem very similar to MSA (especially in short sentences). This is ok if it seems like a more natural translation. Do not try to invent terms to make your translation ``appear dialectal.''

\item Before delivering the translation to the reviewer, carry out your own review first, and make changes when necessary. This will make the entire translation process more efficient for everyone.

\item Use the spelling and writing style you normally use for your dialect. Since there is no single standardized spelling for dialect words, feel free to write them in the way you normally would. For example, in MSA, you can indicate the future tense by using \RL{سـ} at the beginning of a verb or by adding \RL{سوف} (e.g., \RL{سأتصل بك} or \RL{سوف أتصل بك} for ``\emph{I will call you}''). However, in the Egyptian dialect, the future tense is usually marked with \RL{هـ} at the beginning of the verb, like \RL{هكلمك} (``\emph{I will call you}''). Some speakers, however, prefer to use \RL{حـ} instead of \RL{هـ}, so they might write \RL{حكلمك} instead of \RL{هكلمك}.

\end{enumerate}

\section{Annotators' Demographics}
\label{app:demographics}

The annotators' demographics are given in Table~\ref{tab:annotators}.

\begin{table*}[h]
\centering
\small
\begin{tabular}{lllllllll}
\toprule
\textbf{ID} & \textbf{Dataset} & \textbf{Native} & \textbf{Residence} & \textbf{Age} & \textbf{Gender} & \textbf{Degree} & \textbf{Task} & \textbf{Background} \\
\midrule
A0 & MAG & MAG & MAG & 30s & F & BA & A     & RA, LA \\
A1 & MAG & MAG & MAG & 30s & F & BA & T, V  & LA \\
A2 & MAG & MAG & MAG & 40s & M & PhD & T    & LA, RA, CT \\
A3 & MAG & MAG & MAG & 30s & M & MBA & T, V & CT, PT \\
\midrule
A4a & KSA & SYR & SYR & 40s & M & BA& T, V & LA, RA, CT \\
A5 & KSA & SYR & SYR & 40s & F & BA & T, V & LA \\
A6 & KSA & SYR & SYR & 50s & F & BA & A    & LA, PT \\
\midrule
A7 & EGY & EGY & EGY & 30s & F & BA  & T, V & LA, PT \\
A8 & EGY & SYR & SYR & 40s & F & PhD & T, V & CT \\
A9 & EGY & EGY & EGY & 30s & F & BA  & A    & LA, RA, CT \\
\midrule
A4b & SYR & SYR & SYR & 40s & M & BA & A    & LA, RA, CT \\
A10 & SYR & SYR & SYR & 20s & F & BA & T, V & CT, RA \\
A11 & SYR & SYR & SYR & 30s & F & BA & T, V & CT, PT \\
A12 & SYR & SYR & SYR & 40s & F & BA & T    & LA, RA, PT \\
\midrule
A13 & UAE & PAL & UAE & 20s & M & BA & T, V & LA \\
A14 & UAE & PAL & UAE & 40s & M & BA & A & LA \\
A15 & UAE & PAL & UAE & 40s & F & BA & T, V & LA \\
\bottomrule
\end{tabular}

\caption{Annotator demographics and roles for the dialect annotation task. There are three roles: Approve (A), Translate (T), and Validate (V). The annotator background experience includes: Certified Teacher (CT), Private Tutor (PT), Linguistic Annotator (LA), and Research Assistant (RA). All annotators are native speakers of Arabic, and they worked on their own native dialect or on dialects of a country where they had resided for 13--24 years (if they worked on a different dialect).}

\label{tab:annotators}
\end{table*}

\newpage
\section{Additional Similarity Scores}
\label{app:simscores}

We provide additional similarity scores, namely, the weighted Jaccard similarity measure (Table~\ref{tab:app_jacc_w}), and the Manhattan (Table~\ref{tab:app_mann_l1}) and Euclidean (Table~\ref{tab:app_Euc_l2}) distances in support of Sec.~\ref{sec:data_annotation}: similarity between dialects.

\begin{table}[htbp]
\small
\centering
    \begin{tabular}{lrrrrrr}
    \toprule
    \bf Dialect & \bf SYR & \bf UAE & \bf KSA & \bf MSA & \bf MAG & \bf EGY \\    \midrule
    SYR     & 1.000 & 0.554 & 0.601 & 0.533 & 0.442 & 0.593 \\
    UAE     &  & 1.000 & 0.665 & 0.612 & 0.471 & 0.626 \\
    KSA     &  &  & 1.000 & 0.681 & 0.469 & 0.647 \\
    MSA     &  &  &  & 1.000 & 0.455 & 0.601 \\
    MAG     &  &  &  &  & 1.000 & 0.464 \\
    EGY     &  &  &  &  &  & 1.000 \\
    \bottomrule
    \end{tabular}
    \caption{Jaccard weighted similarity.\label{tab:app_jacc_w}}
\end{table}

\begin{table}[htbp]
\small
    \centering
    \begin{tabular}{lrrrrrr}
    \toprule
    \bf Dialect & \bf SYR & \bf UAE & \bf KSA & \bf MSA & \bf MAG & \bf EGY \\
    \midrule
    SYR & 0.000 & 0.713 & 0.630 & 0.776 & 0.988 & 0.649 \\
    UAE &   & 0.000 & 0.518 & 0.625 & 0.936 & 0.594 \\
    KSA &   &   & 0.000 & 0.500 & 0.956 & 0.562 \\
    MSA &   &   &   & 0.000 & 1.000 & 0.661 \\
    MAG &   &   &   &  & 0.000 & 0.972 \\
    EGY &   &   &   &   &  & 0.000  \\
    \bottomrule
    \end{tabular}
    \caption{Manhattan (L1) distance.\label{tab:app_mann_l1}}
\end{table}

\begin{table}[htbh]
\small
    \centering
    \begin{tabular}{lrrrrrr}
    \toprule
    \bf Dialect & \bf SYR & \bf UAE & \bf KSA & \bf MSA & \bf MAG & \bf EGY \\    \midrule
    SYR & 0.000 & 0.715 & 0.648 & 0.756 & 0.758 & 0.745 \\
    UAE &   & 0.000 & 0.506 & 0.653 & 0.843 & 0.448 \\
    KSA &   &   & 0.000 & 0.492 & 0.864 & 0.541 \\
    MSA &   &   &   & 0.000 & 1.000 & 0.648 \\
    MAG &   &   &   &    & 0.000 & 0.874 \\
    EGY &   &   &   &    &      & 0.000  \\
    \bottomrule
    \end{tabular}

\caption{Euclidean (L2) distance.\label{tab:app_Euc_l2}}
\end{table}

\newpage
\section{Dialects and Language Distribution per Domain}\label{app:longStats}

Table~\ref{tab:domain_dialect_questions} gives a detailed statistics of the number of questions for each domain--dialect pair. While the number of questions per topic may vary, the total row shows an equal number of question per dialect/language.

\begin{table}[h]
    \centering
    \begin{tabular}{l|rrrrr|rr}
\toprule
    \bf Domain            & \bf Egyptian & \bf Saudi    & \bf Marrocan & \bf Syrian   & \bf Emirati  & \bf MSA      & \bf English \\
\midrule
abstract\_algebra             & 100      & 100      & 100      & 100      & 100      & 100      & 100      \\
anatomy                       & 100      & 100      & 100      & 100      & 100      & 100      & 100      \\
astronomy                     & 97       & 97       & 97       & 97       & 97       & 97       & 97       \\
business\_ethics              & 99       & 99       & 99       & 99       & 99       & 99       & 99       \\
clinical\_knowledge           & 100      & 100      & 100      & 100      & 100      & 100      & 100      \\
college\_computer\_science    & 100      & 100      & 100      & 100      & 100       & 100      & 100      \\
college\_medicine             & 99       & 99       & 99       & 99       & 99       & 99       & 99       \\
conceptual\_physics           & 100      & 100      & 100      & 100      & 100      & 100      & 100      \\
elementary\_mathematics       & 100      & 100      & 100      & 100      & 100      & 100      & 100      \\
global\_facts                 & 96       & 96       & 96       & 96       & 96       & 96       & 96       \\
high\_school\_chemistry       & 99       & 99       & 99       & 99       & 99       & 99       & 99       \\
high\_school\_geography       & 100      & 100      & 100      & 100      & 100      & 100      & 100      \\
high\_school\_macroeconomics  & 90       & 90       & 90       & 90       & 90       & 90       & 90       \\
high\_school\_psychology      & 100      & 100      & 100      & 100      & 100      & 100      & 100      \\
high\_school\_us\_history     & 100      & 100      & 100      & 100      & 100      & 100      & 100      \\
high\_school\_world\_history  & 100      & 100      & 100      & 100      & 100       & 100      & 100      \\
human\_aging                  & 99       & 99       & 99       & 99       & 99       & 99       & 99       \\
international\_law            & 100      & 100      & 100      & 100      & 100      & 100      & 100      \\
management                    & 100      & 100      & 100      & 100      & 100      & 100      & 100      \\
marketing                     & 68       & 68       & 68       & 68       & 68       & 68       & 68       \\
moral\_scenarios              & 98       & 98       & 98       & 98       & 98       & 98       & 98       \\
nutrition                     & 94       & 94       & 94       & 94       & 94       & 94       & 94       \\
philosophy                    & 100      & 100      & 100      & 100      & 100      & 100      & 100      \\
prehistory                    & 100      & 100      & 100      & 100      & 100      & 100      & 100      \\
professional\_law             & 99       & 99       & 99       & 99       & 99       & 99       & 99       \\
professional\_psychology      & 100      & 100      & 100      & 100      & 100       & 100      & 100      \\
public\_relations             & 98       & 98       & 98       & 98       & 98       & 98       & 98       \\
security\_studies             & 100      & 100      & 100      & 100      & 100      & 100      & 100      \\
sociology                     & 100      & 100      & 100      & 100      & 100      & 100      & 100      \\
us\_foreign\_policy           & 99       & 99       & 99       & 99       & 99       & 99       & 99       \\
virology                      & 100      & 100      & 100      & 100      & 100      & 100      & 100      \\
world\_religions              & 100      & 100      & 100      & 100      & 100      & 100      & 100      \\
\midrule
\textbf{TOTAL}                & 3,135    & 3,135 & 3,135 & 3,135 & 3,135 & 3,135 & 3,135 \\
\bottomrule

    \end{tabular}
    \caption{Number of questions per domain and dialect.}
    \label{tab:domain_dialect_questions}
\end{table}

\newpage

\section{Additional Results}
\label{app:add-results}
In this section, we provide additional results in support of Sec.~\ref{sec:experiments}. In Table~\ref{tab:mmlu_oracle}, we show the accuracy scores for the Oracle \textsc{DialectalArabicMMLU} setting, to compliment the results in Sec.~\ref{subsec:dial_Id}. In Table~\ref{tab:translations_MAD}, we show the change in accuracy when the input is translated using the MADLAD translation model instead of the Google Translate API (Table~\ref{tab:translations_Goog}).

\begin{table*}[h]
    \centering
     \scriptsize
    \tabcolsep7pt
    \begin{tabular}{lrccccc|c|cc}
\toprule
\textbf{Model} & \textbf{Size$\downarrow$} & \textbf{EGY} & \textbf{KSA} & \textbf{MAG} & \textbf{SYR} & \textbf{UAE} & \textbf{DA Avg} & \textbf{MSA} & \textbf{ENG}  \\
\midrule
jais-13b-chat & \textbf{13.0} & 47.0 & 47.1 & 44.1 & 44.6 & 46.9 & 45.9 & 52.0 & 55.3 \\
gemma-3-12b-it & 12.2 & 52.5 & 53.5 & 42.9 & 41.1 & 53.9 & 48.8 & 62.9 & 73.8 \\
Nile-Chat-12B & 11.8 & \textbf{60.3} & \textbf{57.5} & \textbf{54.1} & \textbf{56.6} & \textbf{59.5} & \textbf{57.6} & \textbf{63.9} & 72.8 \\
SILMA-9B-Instruct & 9.2 & 53.2 & 52.7 & 47.6 & 51.5 & 54.4 & 51.9 & 57.6 & 72.4 \\
Fanar-1-9B-Instruct & 8.8 & 55.8 & 55.5 & 51.0 & 53.6 & 56.5 & 54.5 & 61.2 & 70.3 \\
command-r7b-arabic & 8.0 & 52.8 & 52.3 & 48.8 & 51.9 & 53.9 & 51.9 & 57.7 & 67.5 \\
aya-expanse-8b & 8.0 & 50.1 & 49.3 & 46.6 & 48.4 & 49.9 & 48.9 & 53.9 & 63.3 \\
Falcon-H1-7B-Instruct & 7.6 & 55.3 & 55.4 & 51.3 & 52.8 & 57.4 & 54.4 & 62.2 & \textbf{76.4} \\
Mistral-7B-Instruct & 7.2 & 32.4 & 33.8 & 33.6 & 33.7 & 34.9 & 33.7 & 38.1 & 62.7 \\
ALLaM-7B-Instruct & 7.0 & 54.5 & 54.1 & 50.4 & 52.4 & 56.4 & 53.6 & 60.4 & 66.9 \\
Yehia-7B & 7.0 & 52.6 & 53.1 & 48.5 & 52.3 & 53.9 & 52.1 & 58.5 & 62.5 \\
jais-6p7b-chat & 6.8 & 42.5 & 42.7 & 39.1 & 40.3 & 41.6 & 41.2 & 48.2 & 52.9 \\
gemma-3-4b-it & 4.3 & 34.0 & 31.1 & 33.8 & 29.6 & 33.6 & 32.4 & 44.1 & 55.0 \\
Qwen3-4B-Instruct & 4.0 & 32.6 & 28.7 & 27.0 & 28.8 & 30.9 & 29.6 & 31.0 & 65.5 \\
Nile-Chat-4B & 3.9 & 45.8 & 45.6 & 40.3 & 43.9 & 45.6 & 44.2 & 49.5 & 59.5 \\
Falcon-H1-3B-Instruct & 3.1 & 43.3 & 42.5 & 39.6 & 41.6 & 43.9 & 42.2 & 48.3 & 68.0 \\
NileChat-3B & 3.1 & 52.6 & 51.5 & 50.7 & 49.9 & 52.1 & 51.4 & 55.6 & 64.4 \\
jais-2p7b-chat & 2.7 & 38.4 & 40.2 & 34.4 & 37.8 & 39.5 & 38.1 & 43.4 & 47.1 \\
ar-stablelm-2-chat & 1.6 & 36.2 & 36.8 & 35.0 & 34.9 & 36.8 & 35.9 & 37.3 & 38.3 \\
\midrule
\textbf{Average} & 6.8 & 46.9 & 46.5 & 43.1 & 44.5 & 47.5 & 45.7 & 51.9 & 62.9 \\
\bottomrule
\end{tabular}
\caption{Accuracy scores for the Oracle \textsc{DialectalArabicMMLU} setting. (Average of 5 runs for the 32 different topics. Random chance = $\frac{1}{4}$. Size$\downarrow$: Sorted in descending order. \textbf{Bold}: Maximum per column.)\label{tab:mmlu_oracle}}

\end{table*}

\begin{table*}[h]
    \centering
    \scriptsize
    \tabcolsep4pt
    \begin{tabular}{lrrrrr|r||rrrrr|r}
\toprule
&\multicolumn{6}{c||}{\textbf{Dialectal Q\&As translated to English}} &\multicolumn{6}{c}{\textbf{Dialectal Q\&As translated to MSA}* } \\
\textbf{Model}& \textbf{EGY} & \textbf{KSA} & \textbf{MAG} & \textbf{SYR} & \textbf{UAE} & \textbf{DA Avg} & \textbf{EGY} & \textbf{KSA} & \textbf{MAG} & \textbf{SYR} & \textbf{UAE} & \textbf{DA Avg} \\
\midrule
jais-13b-chat & -2.6 & -4.1 & -6.5 & -3.2 & -3.4 & -4.0 & -2.5 & -4.7 & -6.1 & -3.8 & -2.6 & -4.0 \\
gemma-3-12b-it & -6.2 & -5.3 & -8.5 & -5.9 & -5.7 & -6.3 & -6.8 & -6.9 & -10.2 & -6.6 & -8.1 & -7.7 \\
Nile-Chat-12B & -5.9 & -5.9 & -9.2 & -7.4 & -6.0 & -6.8 & -6.9 & -7.9 & -10.7 & -8.7 & -7.7 & -8.4 \\
SILMA-9B-Instruct & -1.0 & -0.7 & -4.2 & -1.1 & 0.0 & -1.4 & -5.0 & -4.9 & -6.1 & -5.2 & -3.5 & -5.0 \\
Fanar-1-9B-Instruct & -5.5 & -4.8 & -10.1 & -5.7 & -5.0 & -6.2 & -6.3 & -6.3 & -10.5 & -6.8 & -5.7 & -7.1 \\
command-r7b-arabic & -1.6 & -4.1 & -7.4 & -5.5 & -2.5 & -4.2 & -4.0 & -4.0 & -8.3 & -5.2 & -4.3 & -5.1 \\
aya-expanse-8b & -2.0 & -1.7 & -6.6 & -2.0 & -0.8 & -2.6 & -3.7 & -3.4 & -7.8 & -4.0 & -3.3 & -4.5 \\
Falcon-H1-7B-Instruct & -0.6 & -2.7 & -5.7 & -1.9 & -1.9 & -2.6 & -6.4 & -7.5 & -9.0 & -7.4 & -7.4 & -7.6 \\
Mistral-7B-Instruct & 14.7 & 11.4 & 6.7 & 12.4 & 11.8 & 11.4 & 0.7 & -1.4 & -2.5 & 0.5 & -0.9 & -0.7 \\
ALLaM-7B-Instruct & -4.2 & -6.3 & -9.9 & -6.0 & -5.9 & -6.5 & -5.1 & -5.6 & -9.8 & -6.5 & -4.9 & -6.5 \\
Yehia-7B & -5.8 & -5.8 & -10.1 & -6.6 & -5.3 & -6.7 & -2.7 & -4.6 & -7.8 & -7.1 & -5.2 & -5.5 \\
jais-6p7b-chat & -0.6 & -2.5 & -3.5 & -1.2 & -1.8 & -1.9 & -1.2 & -3.8 & -4.3 & -1.6 & -2.6 & -2.7 \\
gemma-3-4b-it & 3.2 & 3.8 & 2.7 & 5.3 & 3.0 & 3.7 & -0.6 & 0.1 & -0.7 & 0.3 & -1.0 & -0.4 \\
Qwen3-4B-Instruct & 21.0 & 21.1 & 12.7 & 19.5 & 20.5 & 19.0 & -0.5 & 1.1 & 0.0 & 0.4 & 0.1 & 0.2 \\
Nile-Chat-4B & -3.3 & -2.5 & -3.6 & -1.8 & -1.1 & -2.5 & -5.1 & -6.1 & -5.9 & -4.2 & -4.5 & -5.2 \\
Falcon-H1-3B-Instruct & 4.4 & 4.9 & 1.1 & 3.3 & 5.4 & 3.9 & -4.0 & -2.8 & -5.4 & -3.5 & -4.5 & -4.0 \\
NileChat-3B & -4.1 & -2.5 & -11.6 & -4.4 & -2.7 & -5.1 & -6.0 & -5.3 & -11.8 & -6.1 & -6.1 & -7.0 \\
jais-2p7b-chat & 0.9 & -1.1 & -0.4 & 0.8 & -0.3 & 0.0 & 0.0 & -2.7 & -1.3 & -1.2 & -1.7 & -1.3 \\
ar-stablelm-2-chat & -2.6 & -1.7 & -2.1 & -0.5 & -0.7 & -1.6 & -2.2 & -1.9 & -4.4 & -1.5 & -2.2 & -2.5 \\
\textbf{Average} \tiny (SD) & -0.1  \tiny (7.0)& -0.6 \tiny (6.8)& -4.0  \tiny (6.3)& -0.6 \tiny (6.8) & -0.1 \tiny (6.6)& -1.1 \tiny (6.7)& -3.6 \tiny (2.5)& -4.1 \tiny (2.5)& -6.5 \tiny (3.6)& -4.1\tiny (2.9) & -4.0 \tiny (2.4)& -4.5 \tiny (2.6)\\
\bottomrule
\end{tabular}
\caption{Change in accuracy when using the translated instead of the original inputs. The translation is performed using MADLAD (* The dialectal Arabic questions were translated to English first and then to MSA).}\label{tab:translations_MAD}
\end{table*}

\newpage

\section{Supporting Figures\label{app:B}}
In this section, we provide a set of supporting figures to show the general performance trend of all the evaluated models across the various domains.

\begin{figure*}[htbp]
    \centering
    \includegraphics[width=.75\linewidth]{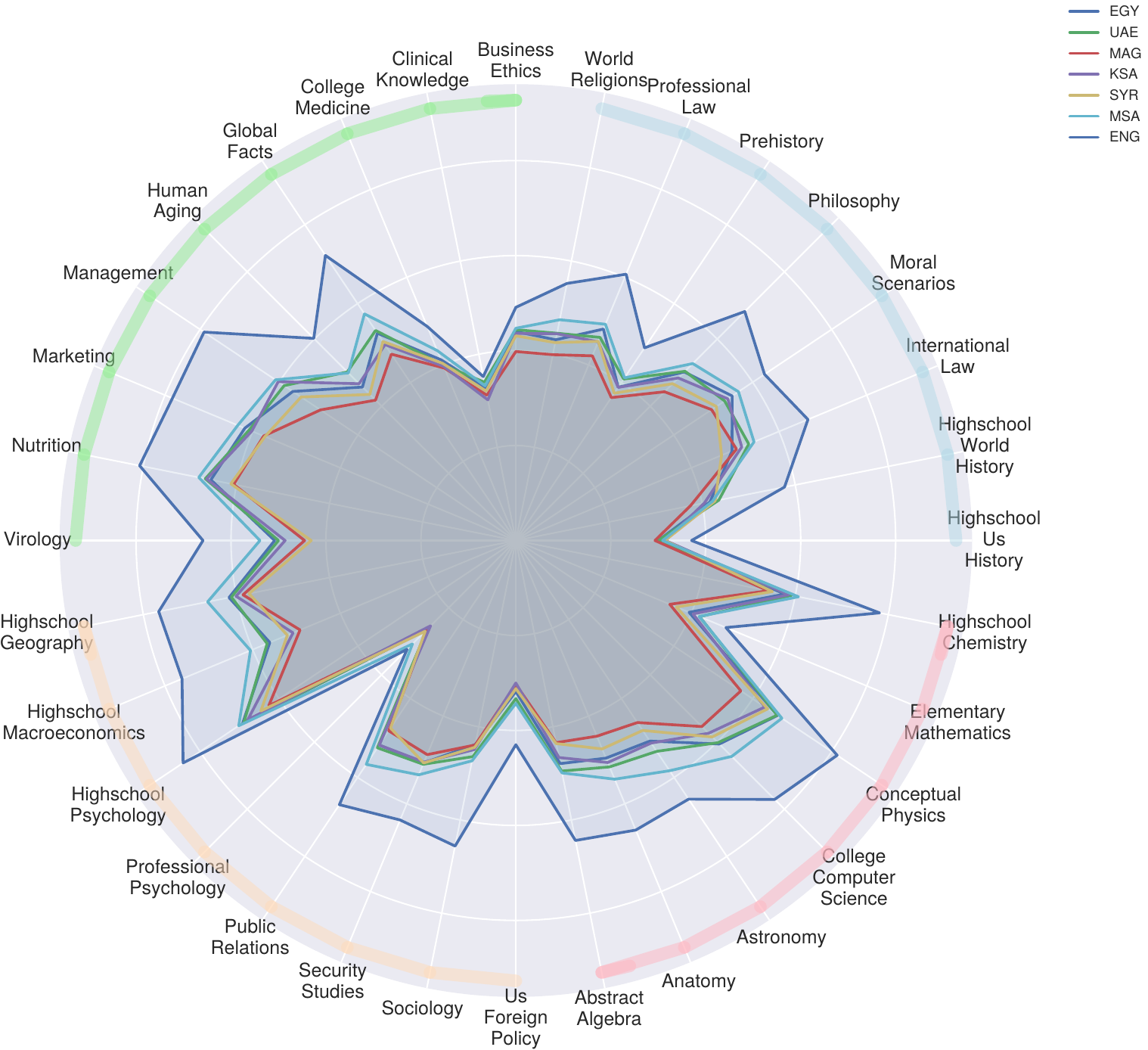}
    \caption{Average \textsc{DialectalArabicMMLU} accuracy per domain (Fields: \textcolor{humanities}{Humanities}, \textcolor{stem}{Stem}, \textcolor{social_sciences}{Social Sciences}, and \textcolor{other}{Other}).}
    \label{fig:dial_mmlu_all}
\end{figure*}

\begin{figure*}[htbp]
    \centering
    \includegraphics[width=.75\linewidth]{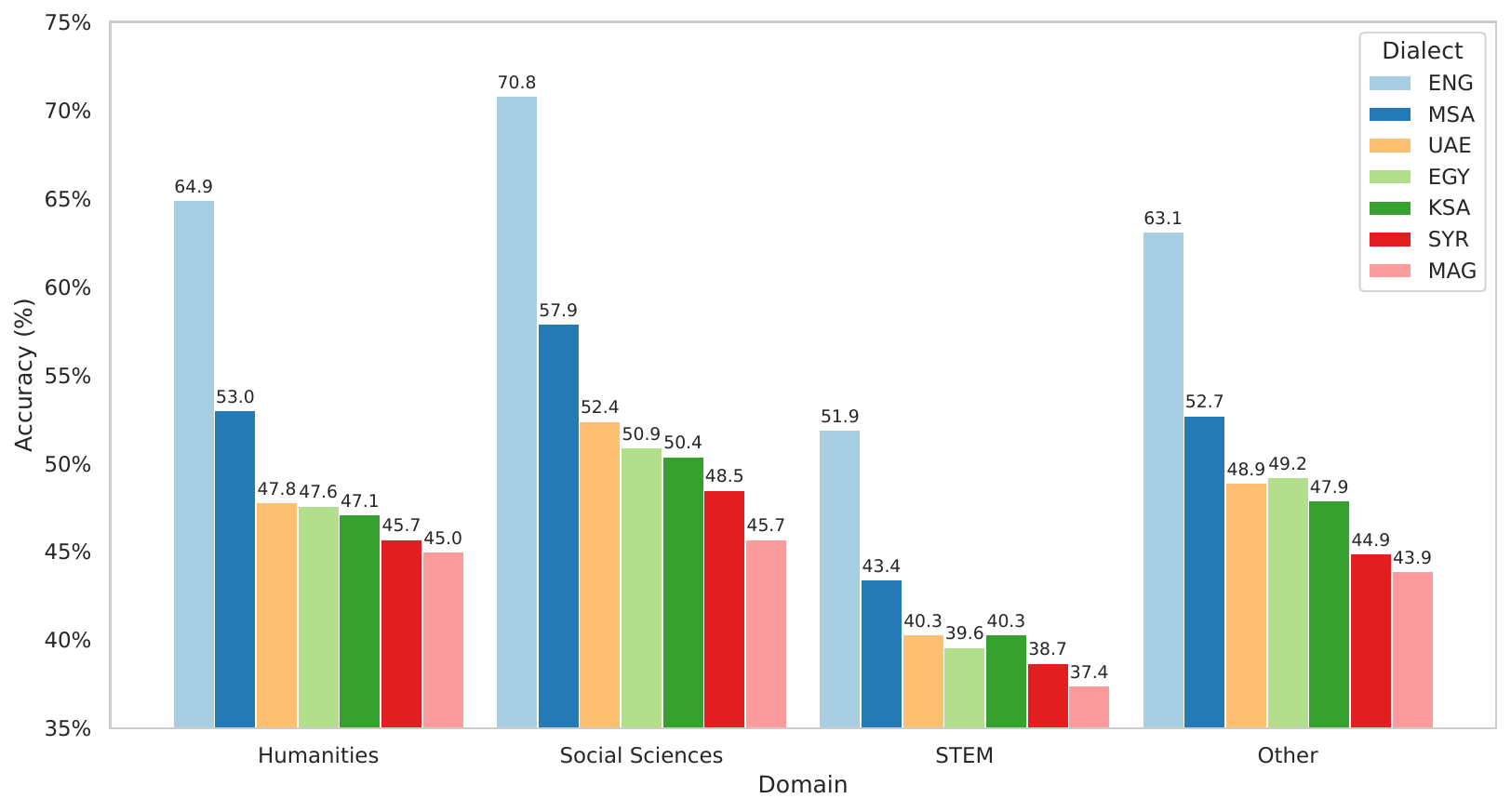}
    \caption{Average \textsc{DialectalArabicMMLU} accuracy grouped by field.}
    \label{fig:dial_mmlu_all_domain}
\end{figure*}

\begin{figure*}[htbp]
    \centering

    \begin{subfigure}[t]{0.3\textwidth}
        \adjustbox{trim= 0 0 0 4, clip, frame}{
            \includegraphics[width=\textwidth]{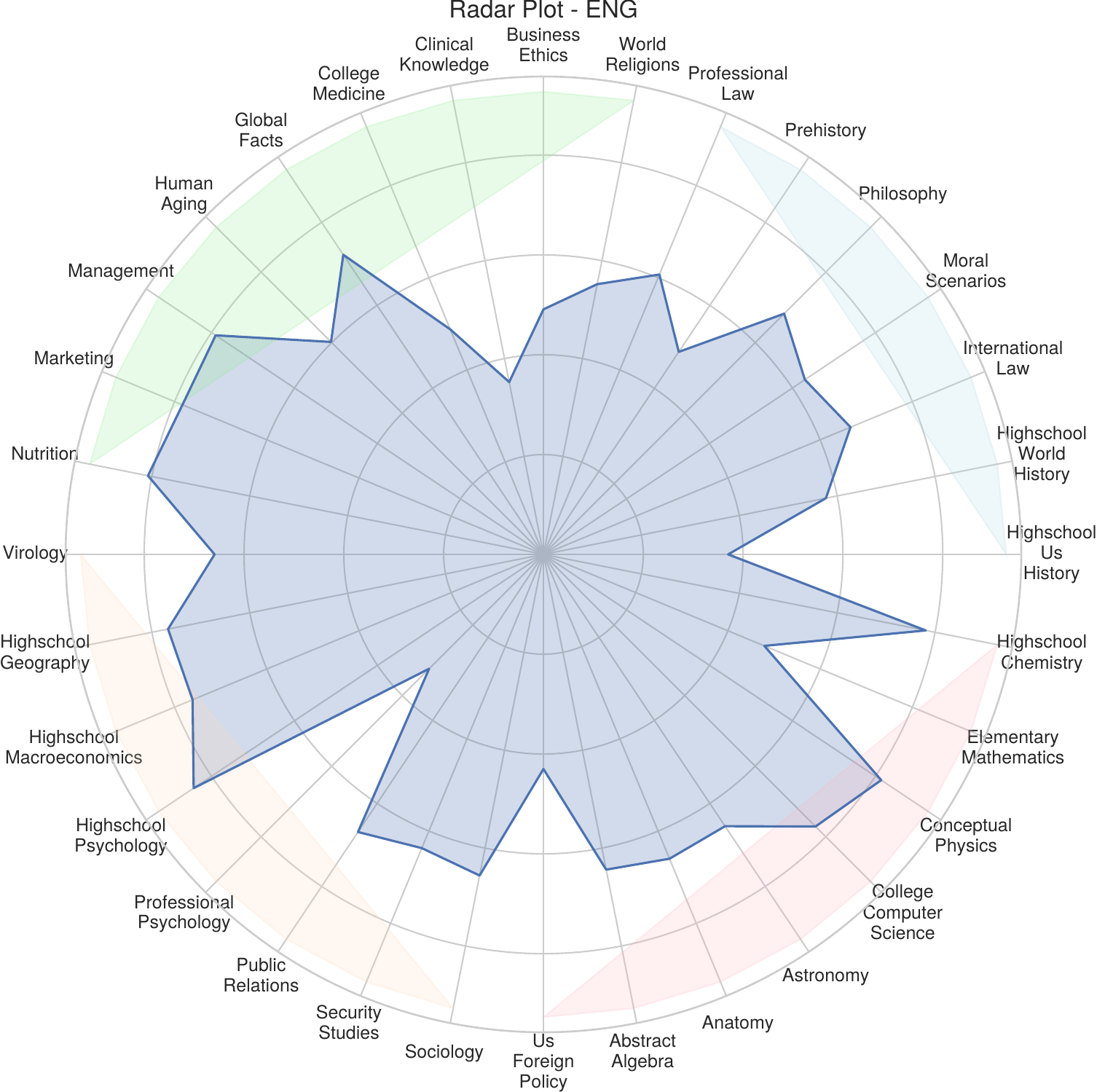}}
        \caption{English}
    \end{subfigure}
    \hfill
    \begin{subfigure}[t]{0.3\textwidth}
        \adjustbox{trim= 0 0 0 4, clip, frame}{
            \includegraphics[width=\textwidth]{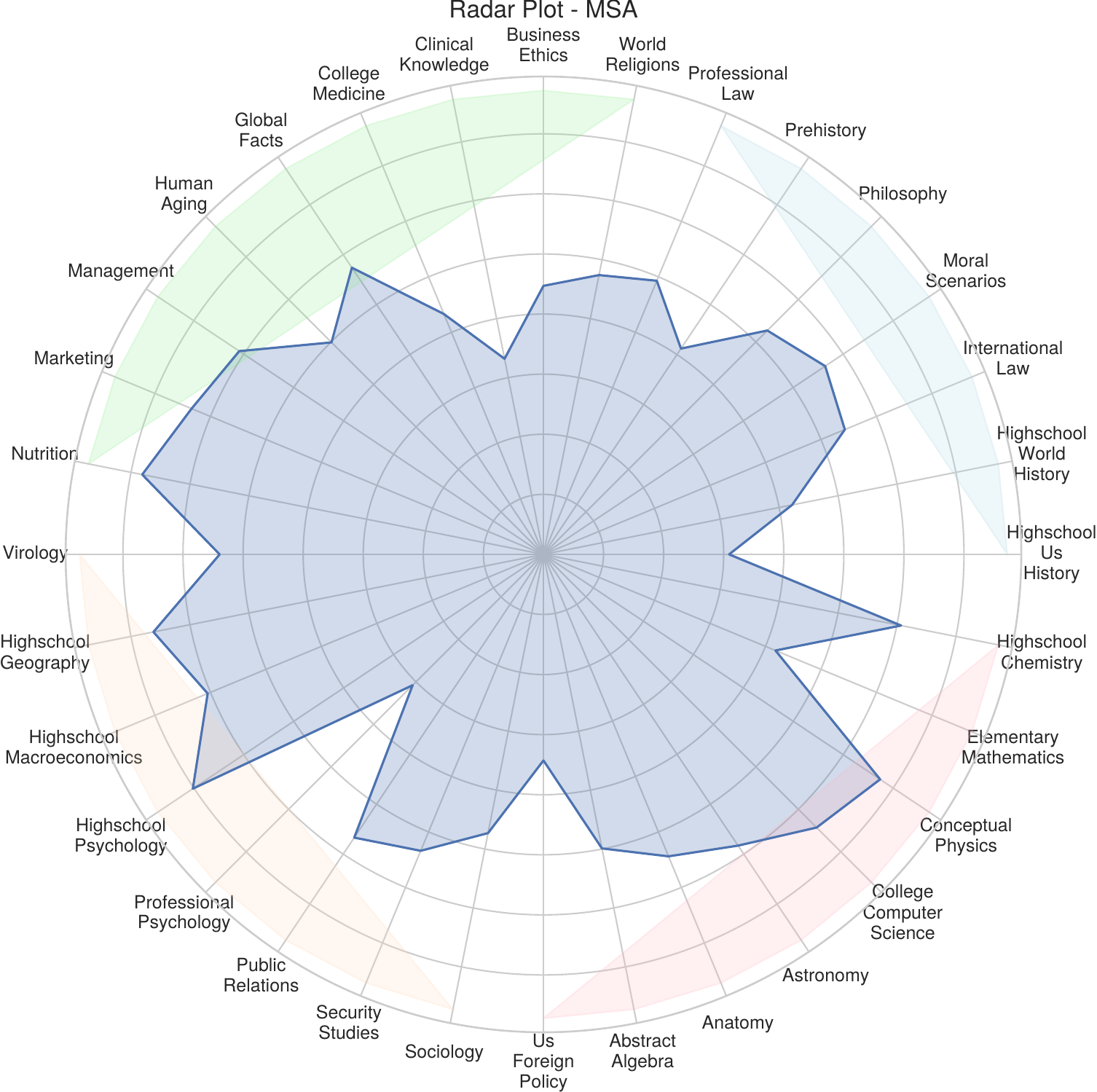}}
        \caption{MSA}
    \end{subfigure}
    \hfill
    \begin{subfigure}[t]{0.3\textwidth}
        \adjustbox{trim=0 0 0 4, clip, frame}{
            \includegraphics[width=\textwidth]{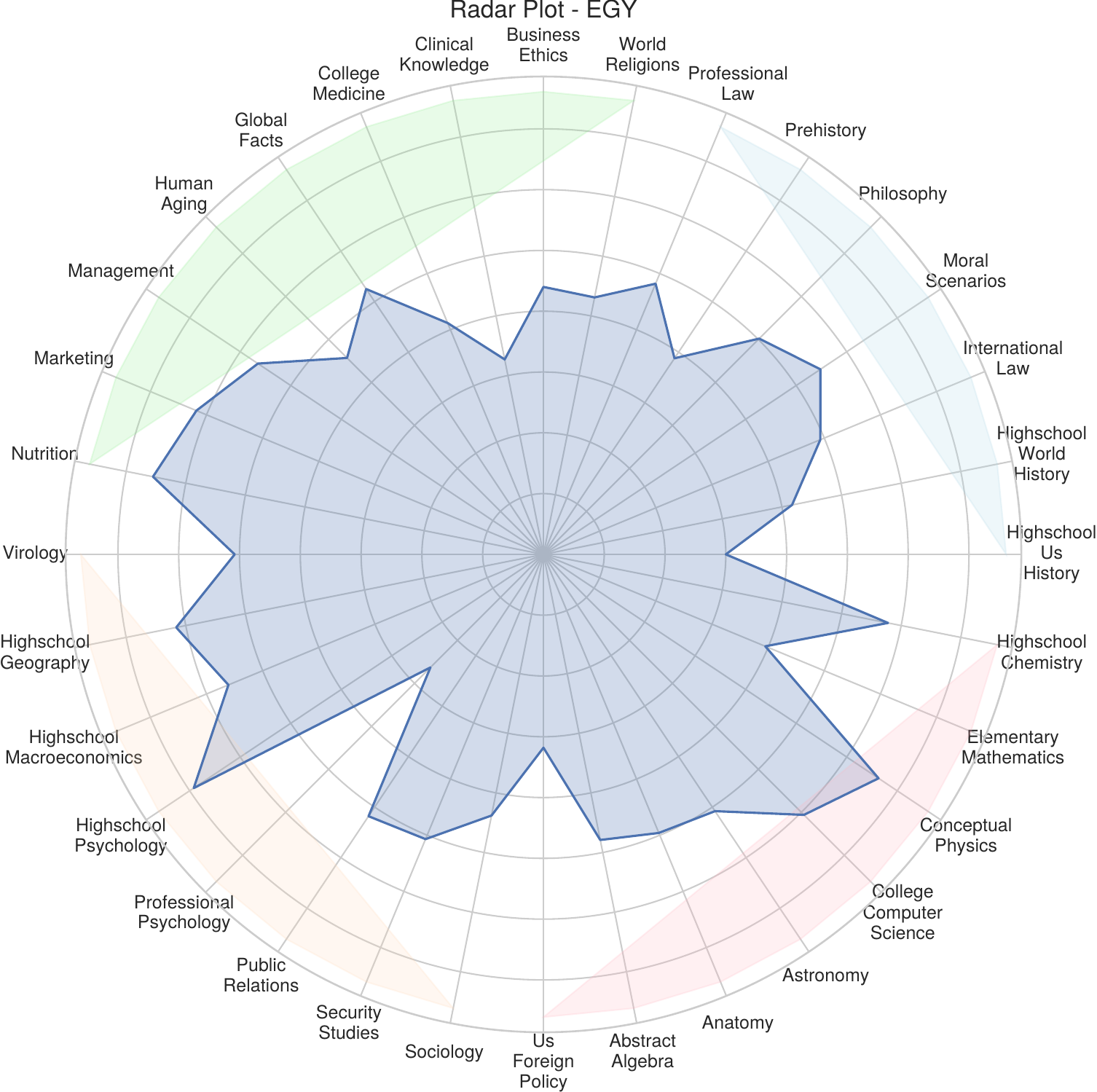}}
        \caption{EGY}
    \end{subfigure}
    \vspace{1em} 


    \begin{subfigure}[t]{0.3\textwidth}
        \adjustbox{trim=0 0 0 4, clip, frame}{
            \includegraphics[width=\textwidth]{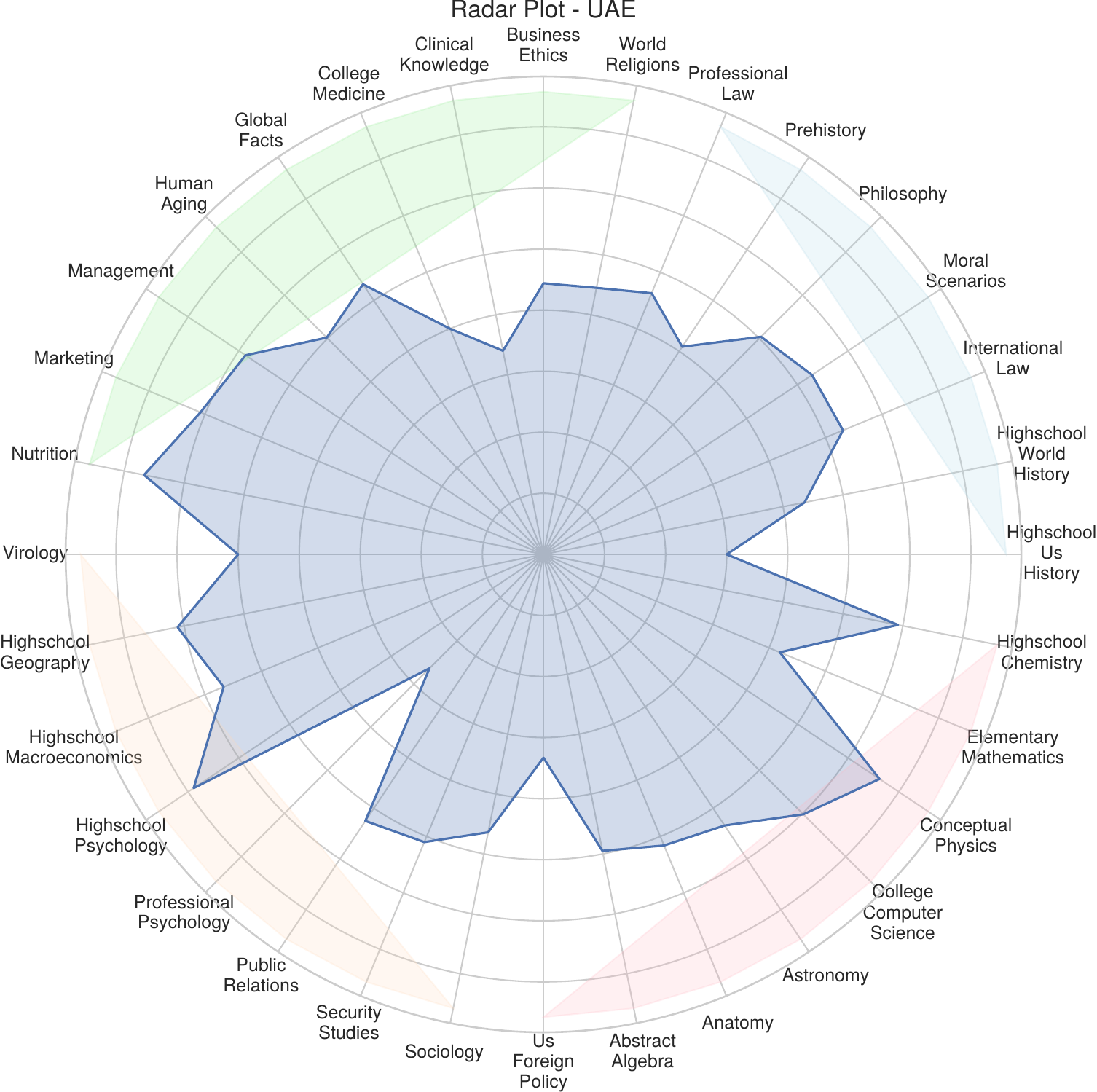}}
        \caption{UAE}
    \end{subfigure}
    \hfill
    \begin{subfigure}[t]{0.3\textwidth}
        \adjustbox{trim=0 0 0 4, clip, frame}{
            \includegraphics[width=\textwidth]{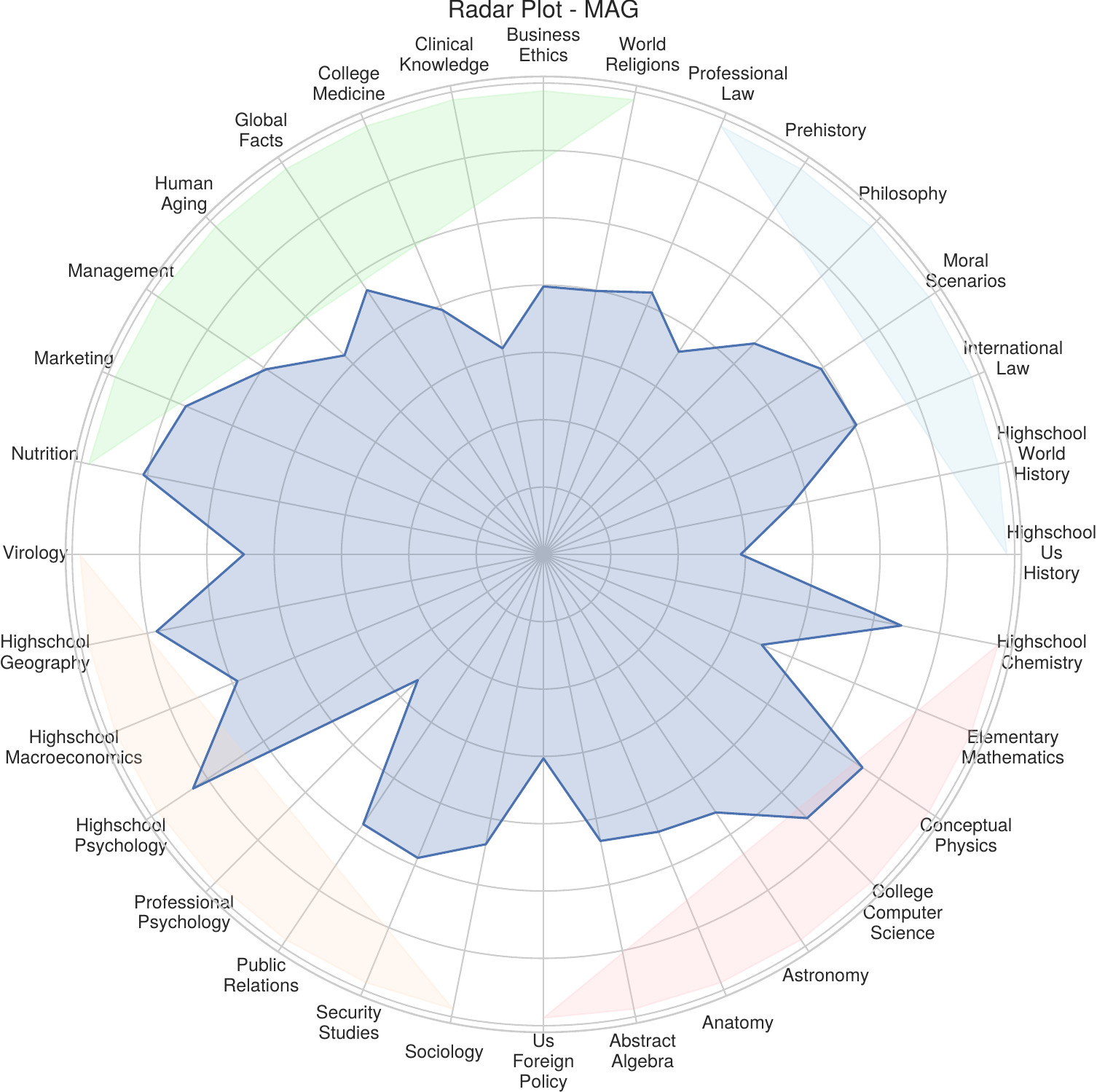}}
        \caption{MAG}
    \end{subfigure}
    \hfill
    \begin{subfigure}[t]{0.3\textwidth}
        \adjustbox{trim=0 0 0 4, clip, frame}{
            \includegraphics[width=\textwidth]{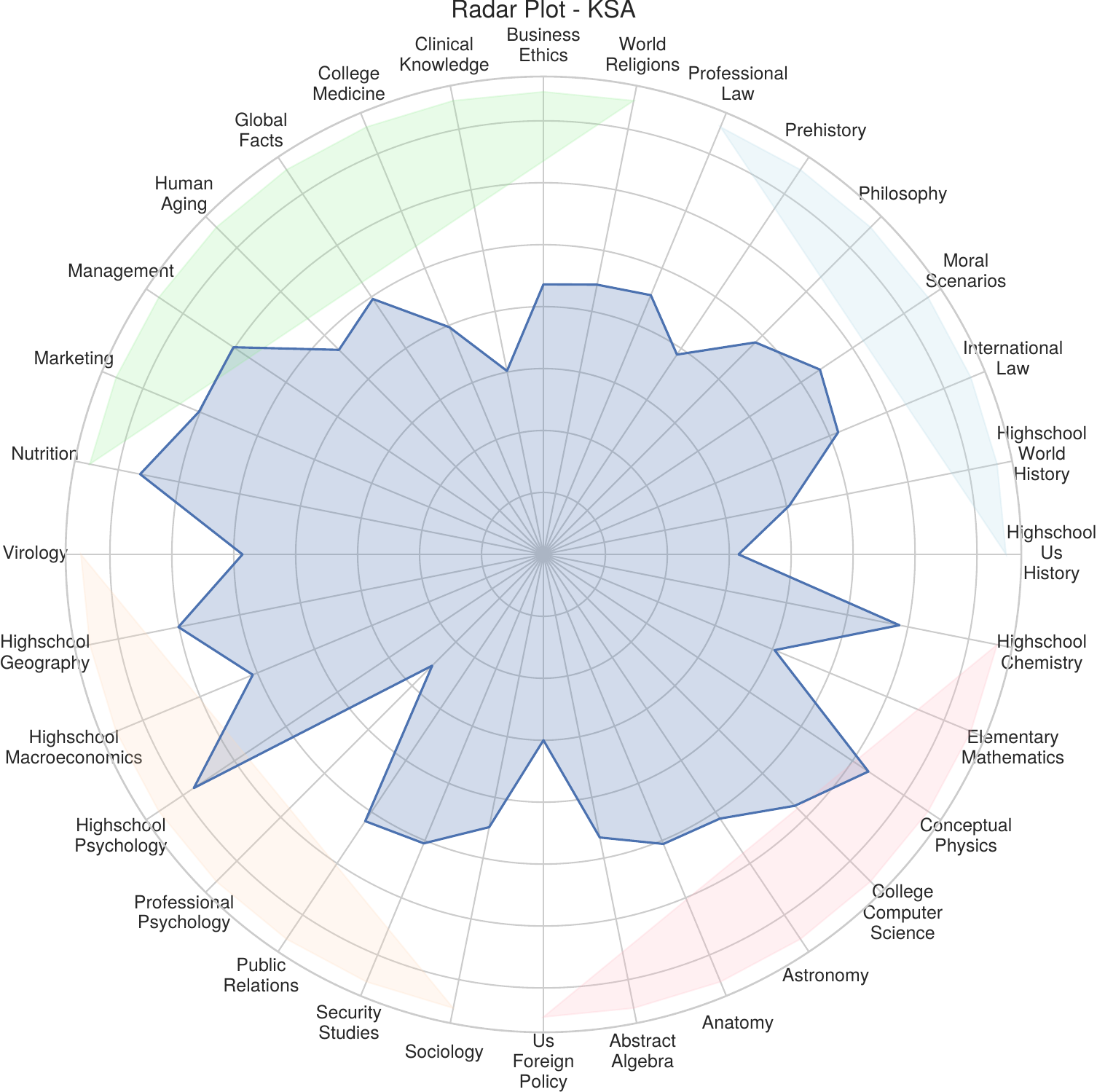}}
        \caption{KSA}
    \end{subfigure}
\vspace{1em}

    \begin{subfigure}[t]{0.3\textwidth}
        \adjustbox{trim=0 0 0 4, clip, frame}{
            \includegraphics[width=\textwidth]{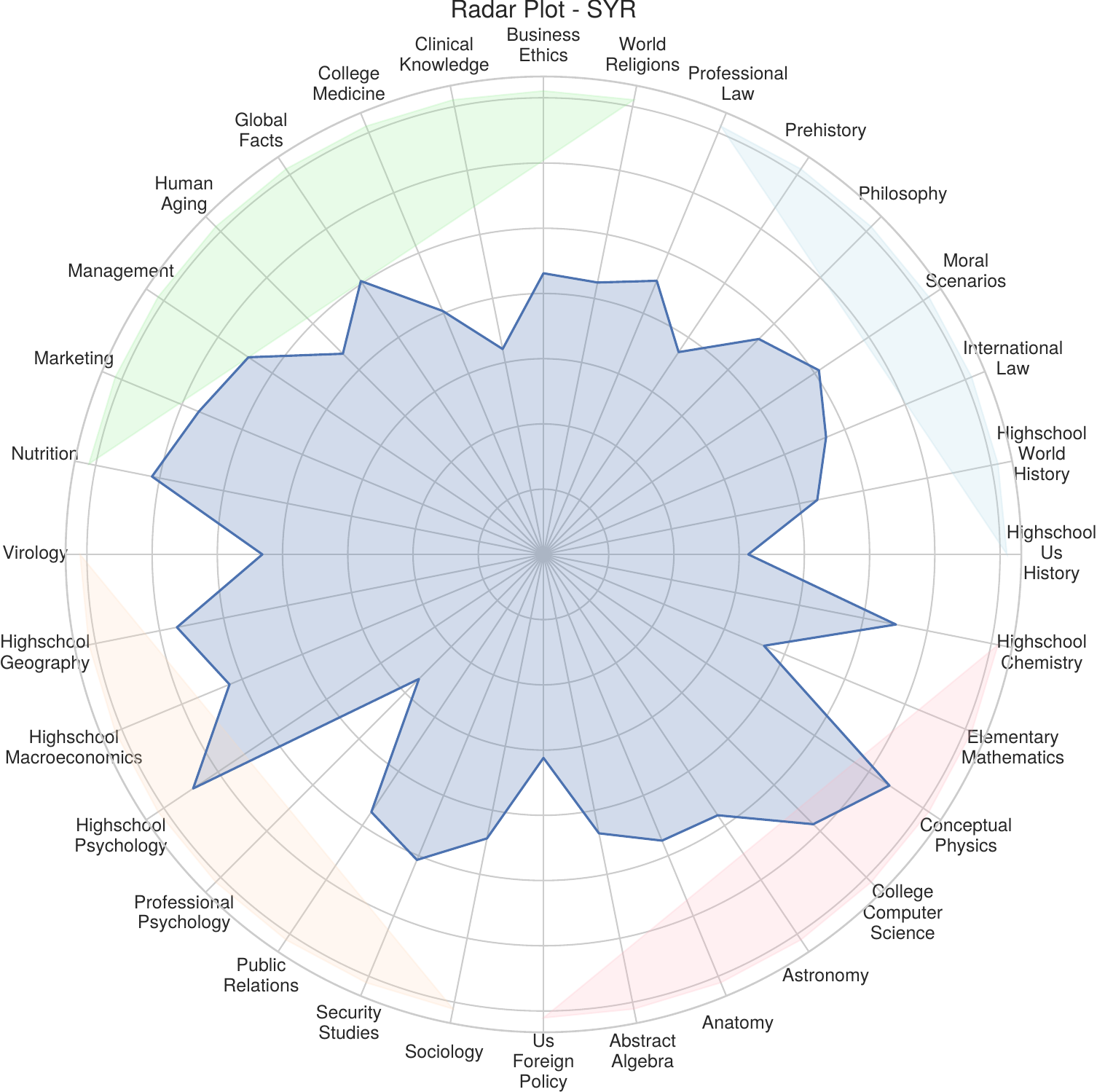}}
        \caption{SYR}
    \end{subfigure}
    \caption{Average \textsc{DialectalArabicMMLU} accuracy over domains for each dialect.
    \label{fig:app_dialmmlu_per_Dial}}
\end{figure*}

\newpage

\clearpage
\section{The Correlation between MCQ and Dialect Identification Accuracy}
\label{app:correlations}
\begin{figure*}[h]
    \centering
    \includegraphics[width=.8\textwidth]{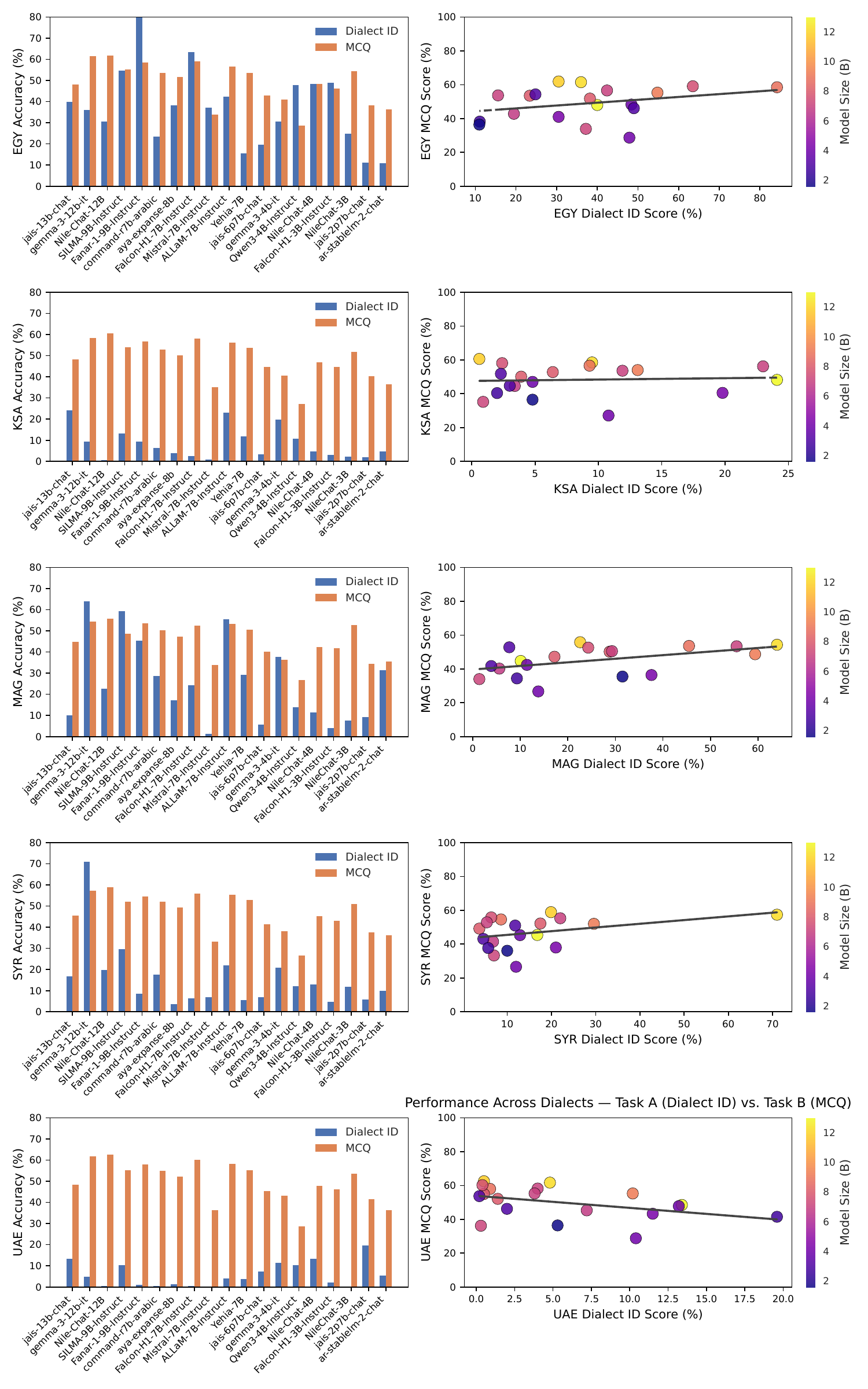}
    \caption{The correlation between MCQ accuracy and dialect identification accuracy per model (left) and on average (right).}
    \label{fig:app_correlations}
\end{figure*}

Fig.~\ref{fig:app_correlations} shows the correlation between each model's performance on the dialect identification task vs. the MCQ task (right) as well as the average performance over all the models (left), for each dialect.  

\end{document}